\documentclass[twoside,11pt]{article}

\usepackage{dnd}
\usepackage{times}

\usepackage[usenames,dvipsnames]{xcolor}
\usepackage[colorlinks,linktoc=all]{hyperref}
\usepackage[all]{hypcap}
\hypersetup{citecolor=BurntOrange}
\hypersetup{linkcolor=MidnightBlue}
\hypersetup{urlcolor=MidnightBlue}

\usepackage{url}
\usepackage{latexsym}
\usepackage{amsmath}
\usepackage{amsfonts}
\usepackage{graphicx}
\usepackage{epstopdf}
\usepackage{multirow}
\usepackage{csquotes}
\usepackage{rotating}
\usepackage{lscape}
\usepackage{gensymb}

\providecommand{\shortcite}[1]{\cite{#1}}

\let\OLDthebibliography\thebibliography
\renewcommand\thebibliography[1]{
  \OLDthebibliography{#1}
  \setlength{\parskip}{0pt}
  \setlength{\itemsep}{0pt plus 0.3ex}
}

\title{A Survey of Available Corpora for Building \\Data-Driven Dialogue Systems}

\author{\name Iulian Vlad Serban \email \{iulian.vlad.serban\} AT umontreal DOT ca \\
  \addr DIRO, Universit{\'e} de Montr{\'e}al \\
  2920 chemin de la Tour, Montr{\'e}al, QC H3C 3J7, Canada\\
  \AND
  \name Ryan Lowe \email \{ryan.lowe\} AT mail DOT mcgill DOT ca\\
  \addr Department of Computer Science, McGill University \\
  3480 University st, Montr{\'e}al, QC H3A 0E9, Canada \\
  \AND
  \name Peter Henderson \email \{peter.henderson\} AT mail DOT mcgill DOT ca\\
  \addr Department of Computer Science, McGill University \\
  3480 University st, Montr{\'e}al, QC H3A 0E9, Canada \\
  \AND
  \name Laurent Charlin \email \{lcharlin\} AT cs DOT mcgill DOT ca \\
  \addr Department of Computer Science, McGill University \\
  3480 University st, Montr{\'e}al, QC H3A 0E9, Canada \\
  \AND
   \name  Joelle Pineau \email \{jpineau\} AT cs DOT mcgill DOT ca \\
  \addr Department of Computer Science, McGill University \\
  3480 University st, Montr{\'e}al, QC H3A 0E9, Canada \\
}

\editor{David Traum}

\date{}

\firstpageno{1}

\begin{document}

\maketitle

\begin{abstract}
During the past decade, several areas of speech and language understanding have witnessed substantial breakthroughs from the use of data-driven models.   In the area of dialogue systems, the trend is less obvious, and most practical systems are still built through significant engineering and expert knowledge.   Nevertheless, several recent results suggest that data-driven approaches are feasible and quite promising.  To facilitate research in this area, we have carried out a wide survey of publicly available datasets suitable for data-driven learning of dialogue systems. We discuss important characteristics of these datasets, how they can be used to learn diverse dialogue strategies, and their other potential uses. We also examine methods for transfer learning between datasets and the use of external knowledge. Finally, we discuss appropriate choice of evaluation metrics for the learning objective.
\end{abstract}


\section{Introduction}
Dialogue systems, also known as interactive conversational agents, virtual agents or sometimes chatterbots, are useful in a wide range of applications ranging from technical support services to language learning tools and entertainment \citep{young2013pomdp,shawar2007chatbots}. 
Large-scale data-driven methods, which use recorded data to automatically infer knowledge and strategies, are becoming increasingly important in speech and language understanding and generation. Speech recognition performance has increased tremendously over the last decade due to innovations in deep learning architectures~\citep{hinton2012deep,Goodfellow-et-al-2015-Book}. Similarly, a wide range of data-driven machine learning methods have been shown to be effective for natural language processing, including tasks relevant to dialogue, such as 
dialogue act classification~\citep{reithinger1997dialogue,stolcke2000dialogue}, dialogue state tracking~\citep{thomson2010bayesian,wang2013simple,ren2013dialog,henderson2013deep,williams2013dialog,henderson2014word,kim2015dialog}, natural language generation ~\citep{langkilde1998generation,oh2000stochastic,walker2002training,ratnaparkhi2002trainable,stent2004trainable,rieser2010natural,mairesse2010phrase,mairesse2014stochastic,wen2015stochastic,sharma2016natural}, and dialogue policy learning~\citep{young2013pomdp}.
We hypothesize that, in general, much of the recent progress is due to the availability of large public datasets, increased computing power, and new machine learning models, such as neural network architectures. To facilitate further research on building data-driven dialogue systems, this paper presents a broad survey of available dialogue corpora. 

Corpus-based learning is not the only approach to training dialogue systems. Researchers have also proposed training dialogue systems online through live interaction with humans, and offline using user simulator models and reinforcement learning methods~\citep{levin1997learning,georgila2006user,paek2006reinforcement,schatzmann2007agenda,jung2009data,schatzmann2009hidden,gavsic2010gaussian,gavsic2011line,daubigney2012comprehensive,gavsic2012policy,su2013dialogue,gasic2013line,pietquin2013survey,young2013pomdp,mohan2014learning,su2015learning,piot2015imitation,cuayahuitl2015strategic,hiraoka2016active,fatemi2016policy,asri2016sequence,williams2016end,su2016continuously}. However, these approaches are beyond the scope of this survey.

This survey is structured as follows.
In the next section, we give a high-level overview of dialogue systems.
We briefly discuss the purpose and goal of dialogue systems.
Then we describe the individual system components that are relevant for data-driven approaches as well as holistic end-to-end dialogue systems.
In Section 3, we discuss types of dialogue interactions and aspects relevant to building data-driven dialogue systems, from a corpus perspective,
as well as modalities recorded in each corpus (e.g.\@ text, speech and video).
We further discuss corpora constructed from both human-human and human-machine interactions, corpora constructed using natural versus unnatural or constrained settings, and corpora constructed using works of fiction.
In Section 4, we present our survey over dialogue corpora according to the categories laid out in Sections 2-3.
In particular, we categorize the corpora based on whether dialogues are between humans or between a human and a machine, and whether the dialogues are in written or spoken language. 
We discuss each corpus in turn while emphasizing how the dialogues were generated and collected, the topic of the dialogues, and the size of the entire corpus. 
In Section 5, we discuss issues related to: corpus size, transfer learning between corpora, incorporation of external knowledge into the dialogue system, data-driven learning for contextualization and personalization, and automatic evaluation metrics.
We conclude the survey in Section 6. 

\section{Characteristics of Data-Driven Dialogue Systems}
This section offers a broad characterization of data-driven dialogue systems, which structures our presentation of the datasets.

\subsection{An Overview of Dialogue Systems}
\begin{figure}[h!t]
        \centering
\includegraphics[trim=0 4cm 0 4cm,clip,width=10cm]{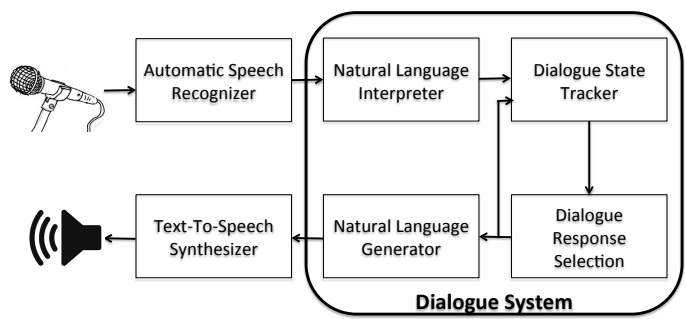}
\caption{Dialogue System Diagram}
\label{fig:dialogue}
\end{figure}

The standard architecture for dialogue systems, shown in Figure~\ref{fig:dialogue}, incorporates a Speech Recognizer, Language Interpreter, State Tracker, Response Generator, Natural Language Generator, and Speech Synthesizer. In the case of text-based (written) dialogues, the Speech Recognizer and Speech Synthesizer can be left out.  While some of the literature on dialogue systems identifies only the State Tracker and Response Selection components as belonging inside the dialogue manager~\citep{young2000probabilistic}, throughout this paper we adopt a broader view where language understanding and generation are incorporated within the dialogue system. This leaves space for the development and analysis of end-to-end dialogue systems~\citep{ritter2011data,vinyals2015neural,lowe2015ubuntu,sordoni2015aneural,shang2015neural,li2015diversity,2015arXiv150704808S,serban2016hierarchical,serban2016multiresolution,dodge2015evaluating,williams2016end,weston2016dialog}.

We focus on corpus-based data-driven dialogue systems. That is, systems composed of machine learning solutions using corpora constructed from real-world data. These system components have variables or parameters that are optimized based on statistics observed in dialogue corpora. In particular, we focus on systems where the majority of variables and parameters are optimized. Such corpus-based data-driven systems should be contrasted to systems where each component is hand-crafted by engineers --- for example, components defined by an a priori fixed set of deterministic rules (e.g.\@ \cite{weizenbaum1966eliza,mcglashan1992dialogue}).
These systems should also be contrasted with systems learning online, such as when the free variables and parameters are optimized directly based on interactions with humans (e.g.\@ \cite{gavsic2011line}).
Still, it is worth noting that it is possible to combine different types of learning within one system. For example, some parameters may be learned using statistics observed in a corpus, while other parameters may be learned through interactions with humans.

While there are substantial opportunities to improve each of the components in Figure~\ref{fig:dialogue} through (corpus-based) data-driven approaches, within this survey we focus primarily on datasets suitable to enhance the components inside the Dialogue System box.
It is worth noting that the Natural Language Interpreter and Generator are core problems in Natural Language Processing with applications well beyond dialogue systems.  

\subsection{Tasks and objectives}
Dialogue systems have been built for a wide range of purposes. A useful distinction can be made between goal-driven dialogue systems, such as technical support services, and non-goal-driven dialogue systems, such as language learning tools or computer game characters. Although both types of systems do in fact have objectives, typically the goal-driven dialogue systems have a well-defined measure of performance that is explicitly related to task completion.

\textbf{Non-goal-driven Dialogue Systems}.
Research on non-goal-driven dialogue systems goes back to the mid-60s. It began, perhaps, with Weizenbaum's famous program \textit{ELIZA}, a system based only on simple text parsing rules that managed to convincingly mimic a Rogerian psychotherapist by persistently rephrasing statements or asking questions \citep{weizenbaum1966eliza}.
This line of research was continued by \shortcite{BBS:2533884}, who used simple text parsing rules to construct the dialogue system \textit{PARRY}, which managed to mimic the pathological behaviour of a paranoid patient to the extent that clinicians could not distinguish it from real patients.
However, neither of these two systems used data-driven learning approaches.
Later work, such as the MegaHal system by \shortcite{Hutchens:1998:IM:1603899.1603945}, started to apply data-driven methods \citep{shawar2007chatbots}.
\shortcite{Hutchens:1998:IM:1603899.1603945} proposed modelling dialogue as a stochastic sequence of discrete symbols (words) using $4$'th order Markov chains.
Given a user utterance, their system generated a response by following a two-step procedure:
first, a sequence of topic keywords, used to create a seed reply, was extracted from the user's utterance; second, starting from the seed reply, two separate Markov chains generated the words preceding and proceeding the seed keywords.
This procedure produced many candidate responses, from which the highest entropy response was returned to the user.
Under the assumption that the coverage of different topics and general fluency is of primary importance, the $4$'th order Markov chains were trained on a mixture of data sources ranging from real and fictive dialogues to arbitrary texts.
Unfortunately, until very recently, such data-driven dialogue systems were not applied widely in real-world applications~\citep{perez2011conversational,shawar2007chatbots}.
Part of the reason for this might be due to their non-goal-driven nature, which made them hard to commercialize.
Another barrier to commercialization might have been the lack of theoretical and empirical understanding of such systems.
Nevertheless, in a similar spirit over the past few years, neural network architectures trained on large-scale corpora have been investigated.
These models have demonstrated promising results for several non-goal-driven dialogue tasks~\citep{ritter2011data,vinyals2015neural,lowe2015ubuntu,sordoni2015aneural,shang2015neural,li2015diversity,2015arXiv150704808S,serban2016hierarchical,serban2016multiresolution,dodge2015evaluating,williams2016end,weston2016dialog}.
However, they require having sufficiently large corpora --- in the hundreds of millions or even billions of words --- in order to achieve these results.

\textbf{Goal-driven Dialogue Systems}.
Initial work on goal-driven dialogue systems was primarily based on deterministic hand-crafted rules coupled with learned speech recognition models (e.g.\@ off-the-shelf speech recognition software).
One example is the SUNDIAL project, which was capable of providing timetable information about trains and airplanes, as well as taking airplane reservations~\citep{aust1995philips,mcglashan1992dialogue,simpson1993black}.
Later, machine learning techniques were used to classify the intention (or need) of the user, as well as to bridge the gap between text and speech (e.g.\@ by taking into account uncertainty related to the outputs of the speech recognition model)~\citep{gorin1997may}.
Research in this area started to take off during the mid 1990s, when researchers began to formulate dialogue as a sequential decision making problem based on Markov decision processes \citep{singh1999reinforcement,young2013pomdp,paek2006reinforcement,pieraccini2009we}. 
Unlike non-goal-driven systems, industry played a major role and enabled researchers to have access to (at the time) relatively large dialogue corpora for certain tasks, such as recordings from technical support call centres. Although research in the past decade has continued to push the field towards data-driven approaches, commercial systems are highly domain-specific and heavily based on hand-crafted rules and features \citep{young2013pomdp}.
In particular, many of the tasks and datasets available are constrained to narrow domains.

\subsection{Learning Dialogue System Components} \label{subsec:discriminative_models}
Modern dialogue systems consist of several components, as illustrated in Figure~\ref{fig:dialogue}.
Several of the dialogue system components can be learned through so-called discriminative models, which aim to predict labels or annotations relevant to other parts of the dialogue system. 
Discriminative models fall into the machine learning paradigm of supervised learning. When the labels of interest are discrete, the models are called \emph{classification} models, which is the most common case.
When the labels of interest are continuous, the models are called \emph{regression} models.
One popular approach for tackling the discriminative task is to learn a probabilistic model of the labels conditioned on the available information $P(Y | X)$, where $Y$ is the label of interest (e.g.\@ a discrete variable representing the user intent) and $X$ is the available information (e.g.\@ utterances in the conversation).
Another popular approach is to use maximum margin classifiers, such as support vector machines \citep{Cristianini1999}.

Although it is beyond the scope of this paper to provide a survey over such system components, we now give a brief example of each component. This will motivate and facilitate the dataset analysis.

\textbf{Natural Language Interpreter}.
An example of a discriminative model is the user intent classification model, which acts as the Natural Language Interpreter.
This model is trained to predict the intent of a user conditioned on the utterances of that user.
In this case, the intent is called the \emph{label} (or \emph{target} or \emph{output}), and the conditioned utterances are called the \emph{conditioning variables} (or \emph{inputs}).
Training this model requires examples of pairs of user utterances and intentions.
One way to obtain these example pairs would be to first record written dialogues between humans carrying out a task, and then to have humans annotate each utterance with its intention label.
Depending on the complexity of the domain, this may require training the human annotators to reach a certain level of agreement between annotators.

\textbf{Dialogue State Tracker}.
A Dialogue State Tracker might similarly be implemented as a classification model \citep{williams2013dialog}.
At any given point in the dialogue, such a model will take as input all the user utterances and user intention labels estimated by a Natural Language Interpreter model so far and output a distribution over possible dialogue states.
One common way to represent dialogue states are through slot-value pairs.
For example, a dialogue system providing timetable information for trains might have three different slots: \textit{departure city, arrival city, and departure time}.
Each slot may take one of several discrete values (e.g.\@ \textit{departure city} could take values from a list of city names).
The task of the Dialogue State Tracker is then to output a distribution over every possible combination of slot-value pairs.
This distribution --- or alternatively, the $K$ dialogue states with the highest probability --- may then be used by other parts of the dialogue system.
The Dialogue State Tracker model can be trained on examples of dialogue utterances and dialogue states labelled by humans.

\textbf{Dialogue Response Selection}.
Given the dialogue state distribution provided by the Dialogue State Tracker, the Dialogue Response Selection component must select the correct system response (or action).
This component may also be implemented as a classification model that maps dialogue states to a probability over a discrete set of responses.
For example, in a dialogue system providing timetable information for trains, the set of responses might include \textit{providing information} (e.g.\@ providing the departure time of the next train with a specific departure and arrival city) and \textit{clarification questions} (e.g.\@ asking the user to re-state their departure city).
The model may be trained on example pairs of dialogue states and responses.

\textbf{Natural Language Generator}.
Given a dialogue system response (e.g.\@ a response providing the departure time of a train), the Natural Language Generator must output the natural language utterance of the system.
This has often been implemented in commercial goal-driven dialogue systems using hand-crafted rules.
Another option is to learn a discriminative model to select a natural language response.
In this case, the output space may be defined as a set of so-called \textit{surface form} sentences (e.g.\@ \textit{"The requested train leaves city X at time Y"}, where X and Y are placeholder values).
Given the system response, the classification model must choose an appropriate surface form.
Afterwards, the chosen surface form will have the placeholder values substituted in appropriately (e.g.\@ X will be replaced by the appropriate city name through a database look up).
As with other classification models, this model may be trained on example pairs of system responses and surface forms.

Discriminative models have allowed goal-driven dialogue systems to make significant progress \citep{williams2013dialog}.
With proper annotations, discriminative models can be evaluated automatically and accurately.
Furthermore, once trained on a given dataset, these models may be plugged into a fully-deployed dialogue system (e.g.\@ a classification model for user intents may be used as input to a dialogue state tracker). 

\subsection{End-to-end Dialogue Systems}\label{sec:DiscVsGen}
Not all dialogue systems conform to the architecture shown in Figure~\ref{fig:dialogue}.
In particular, so-called end-to-end dialogue system architectures based on neural networks have shown promising results on several dialogue tasks~\citep{ritter2011data,vinyals2015neural,lowe2015ubuntu,sordoni2015aneural,shang2015neural,li2015diversity,2015arXiv150704808S,serban2016hierarchical,serban2016multiresolution,dodge2015evaluating}.
In their purest form, these models take as input a dialogue in text form and output a response (or a distribution over responses).
We call these systems end-to-end dialogue systems because they possess two important properties.
First, they do not contain or require learning any sub-components (such as Natural Language Interpreters or Dialogue State Trackers). Consequently, there is no need to collect intermediate labels (e.g.\@ user intention or dialogue state labels).
Second, all model parameters are optimized w.r.t.\@ a single objective function.
Often the objective function chosen is maximum log-likelihood (or cross-entropy) on a fixed corpus of dialogues.
Although in the original formulation these models depended only on the dialogue context, they may be extended to also depend on outputs from other components (e.g.\@ outputs from the speech recognition tracker), and on external knowledge (e.g.\@ external databases).

End-to-end dialogue systems can be divided into two categories: those that select deterministically from a fixed set of possible responses, and those that attempt to generate responses by keeping a posterior distribution over possible utterances.
Systems in the first category map the dialogue history, tracker outputs and external knowledge (e.g.\@ a database, which can be queried by the system) to a response action:
\begin{align}
f_{\theta} \ : & \ \{ \text{dialogue history},\; \text{tracker outputs}, \text{external knowledge} \}
 \to \ \text{action} \ a_t, \label{eq:deterministic_model}
\end{align}
where $a_t$ is the dialogue system response action at time $t$, and $\theta$ is the set of parameters that defines $f$.
Information retrieval and ranking-based systems --- systems that search through a database of dialogues and pick responses with the most similar context, such as the model proposed by \shortcite{banchs2012iris} --- belong to this category.
In this case, the mapping function $f_{\theta}$  projects the dialogue history into a Euclidean space (e.g. using TF-IDF bag-of-words representations). The response is then found by projecting all potential responses into the same Euclidean space, and the response closest to the desirable response region is selected.
The neural network proposed by \shortcite{lowe2015ubuntu} also belongs to this category.
In this case, the dialogue history is projected into a Euclidean space using a recurrent neural network encoding the dialogue word-by-word. 
Similarly, a set of candidate responses are mapped into the same Euclidean space using another recurrent neural network encoding the response word-by-word.
Finally, a relevance score is computed between the dialogue context and each candidate response, and the response with the highest score is returned.
Hybrid or combined models, such as the model built on both a phrase-based statistical machine translation system and a recurrent neural network proposed by \shortcite{sordoni2015aneural}, also belong to this category. In this case, a response is generated by deterministically creating a fixed number of answers using the machine translation system and then picking the response according to the score given by a a neural network. Although both of its sub-components are based on probabilistic models, the final model does not construct a probability distribution over all possible responses.\footnote{Although the model does not require intermediate labels, it consists of sub-components whose parameters are trained with different objective functions. Therefore, strictly speaking, this is not an end-to-end model.}

In contrast to a deterministic system, a generative system explicitly computes a full posterior probability distribution over possible system response actions at every turn:
\begin{align}
P_{\theta}(\text{action $a_t$} \  & | \ \text{dialogue history}, \text{tracker outputs}, 
 \text{external knowledge}). 
 \label{eq:generative_model}
\end{align}
Systems based on generative recurrent neural networks belong to this category \citep{vinyals2015neural}. By breaking down eq.\@ \eqref{eq:generative_model} into a product of probabilities over words, responses can be generated by sampling word-by-word from their probability distribution. Unlike the deterministic response models, these systems are also able to generate entirely novel responses (e.g.\@ by sampling word-by-word). Highly probable responses, i.e.\@ the response with the highest probability, can further be generated by using a method known as beam-search \citep{graves2012seq}. These systems project each word into a Euclidean space (known as a word embedding) \citep{bengio2003neural}; they also project the dialogue history and external knowledge into a Euclidean space \citep{wen2015stochastic,lowe2015incorporating}. 
Similarly, the system proposed by \shortcite{ritter2011data} belongs to this category. 
Their model uses a statistical machine translation model to map a dialogue history to its response.
When trained solely on text, these generative models can be viewed as unsupervised learning models, because they aim to reproduce data distributions. In other words, the models learn to assign a probability to every possible conversation, and since they generate responses word by word, they must learn to simulate the behaviour of the agents in the training corpus.

Early reinforcement learning dialogue systems with stochastic policies also belong to this category (the NJFun system \citep{singh2002optimizing} is an example of this).
In contrast to the neural network and statistical machine translation systems, these reinforcement learning systems typically have very small sets of possible hand-crafted system states (e.g.\@ hand-crafted features describing the dialogue state).
The action space is also limited to a small set of pre-defined responses.
This makes it possible to apply established reinforcement learning algorithms to train them either online or offline, however it also severely limits their application area.
As Singh et al.\@ \citep[p.5]{singh2002optimizing} remark: ``We view the design of an appropriate state space as application-dependent, and a task for a skilled system designer.''

\section{Dialogue Interaction Types \& Aspects}
This section provides a high-level discussion of different types of dialogue interactions and their salient aspects. The categorization of dialogues is useful for understanding the utility of various datasets for particular applications, as well as for grouping these datasets together to demonstrate available corpora in a given area.

\subsection{Written, Spoken \& Multi-modal Corpora} \label{subsec:spoken_versus_written}

An important distinction between dialogue corpora is whether participants (interlocutors) interact through written language, spoken language, or in a multi-modal setting (e.g.\@ using both speech and visual modalities).
Written and spoken language differ substantially w.r.t.\@ their linguistic properties.
.
Spoken language tends to be less formal, containing lower information content and many more pronouns than written language \citep{carter2006cambridge,biber2001diachronic,biber1986initial}.
In particular, the differences are magnified when written language is compared to spoken face-to-face conversations, which are multi-modal and highly socially situated.
As \shortcite{biber1986initial} observed, 
pronouns, questions, and contradictions, as well as that-clauses and if-clauses, appear with a high frequency in face-to-face conversations. \shortcite{forchini2012movie} summarized these differences: \textit{``... studies show that face-to-face conversation is interpersonal, situation-dependent has no narrative concern or as Biber and Finegan (1986) put it, is a highly interactive, situated and immediate text type...''}
Due to these differences between spoken and written language, we will emphasize the distinction between dialogue corpora in written and spoken language in the following sections.

Similarly, dialogues involving visual and other modalities differ from dialogues without these modalities \citep{Card:1983:PHI:578027,Goodwin:1981:Conversational}.
When a visual modality is available --- for example, when two human interlucators converse face-to-face --- body language and eye gaze has a significant impact on what is said and how it is said \citep{gibson1963perception,lord1974perception,cooper1974control,chartrand1999chameleon,de2013speaker}.
Aside from the visual modality, dialogue systems may also incorporate other situational modalities, including aspects of virtual environments \citep{rickel1999animated,traum2002embodied} and user profiles \citep{li2016persona}.







\subsection{Human-Human Vs.\@ Human-Machine Corpora} \label{subsec:humanhuman_vs_humansystem_corpora}
Another important distinction between dialogue datasets resides in the types of interlocutors --- notably, whether it involves interactions between two humans, or between a human and a computer\footnote{Machine-machine dialogue corpora are not of interest to us, because they typically differ significantly from natural human language. Furthermore, user simulation models are outside the scope of this survey.}. The distinction is important because current artificial dialogue systems are significantly constrained.
These systems do not produce nearly the same distribution of possible responses as humans do under equivalent circumstances.
As stated by~\shortcite{williams2007partially}: 
\begin{displayquote}
(Human-human conversation) does not contain the same distribution of understanding errors, and human--human turn-taking is much richer than human-machine dialog. As a result, human-machine dialogue exhibits very different traits than human-human dialogue (Doran et al., 2001; Moore and Browning, 1992).
\end{displayquote}
The expectation a human interlucator begins with, and the interface through which they interact, also affect the nature of the conversation \citep{jonsson1988talking}.

For goal-driven settings, \shortcite{williams2007partially} have previously argued against building data-driven dialogue systems using human-human dialogues: \textit{``... using human-human conversation data is not appropriate because it does not contain the same distribution of understanding errors, and because human-human turn-taking is much richer than human-machine dialog.''}
This line of reasoning seems particularly applicable to spoken dialogue systems, where speech recognition errors can have a critical impact on performance and therefore must be taken into account when learning the dialogue model.
The argument is also relevant to goal-driven dialogue systems, where an effective dialogue model can often be learned using reinforcement learning techniques. \shortcite{williams2007partially} also argue against learning from corpora generated between humans and existing dialogue systems: \textit{``While it would be possible to use a corpus collected from an existing spoken dialogue system, supervised learning would simply learn to approximate the policy used by that spoken dialogue system and an overall performance improvement would therefore be unlikely.''}

Thus, it appears, for goal-driven spoken dialogue systems in particular, that the most effective strategy is learning online through interaction with real users.
Nonetheless, there exists useful human-machine corpora where the interacting machine uses a stochastic policy that can generate sufficient coverage of the task (e.g.\@ enough \textit{good} and enough \textit{bad} dialogue examples) to allow an effective dialogue model to be learned. In this case, the goal is to learn a policy that is eventually better than the original stochastic policy used to generate the corpus through a process known as bootstrapping.

In this survey we focus on data-driven learning from human-human and human-machine dialogue corpora.
Despite the advantages of learning online through interactions with real users, learning based on human-human dialogue corpora may be more suitable for open domain dialogue systems because they reflect natural dialogue interactions.
By natural dialogues, we mean conversations that are unconstrained and unscripted, e.g.\@ between interlocutors who are not instructed to carry out a particular task, to follow a series of instructions, or to act out a scripted dialogue.
In this setting, the dialogue process is relatively unaffected by researchers, e.g.\@ the interlocutors are not interrupted by question prompts in the middle of a dialogue.
As can be expected, such conversations include a significant amount of turn-taking, pauses and common grounding phenomena \citep{clark1991grounding}.
Additionally, they are more diverse, and open up the possibility for the model to learn to understand natural language.

\subsection{Natural Vs.\@ Unnatural Corpora} \label{subsec:natural_vs_unnatural_corpora}
The way in which a dialogue corpus is generated and collected can have a significant influence on the trained data-driven dialogue system. In the case of human-human dialogues, an ideal corpus should closely resemble natural dialogues between humans.
Arguably, this is the case when conversations between humans are recorded and transcribed, and when the humans in the dialogue represent the true population of users with whom the dialogue system is intended to interact. It is even better if they are unaware of the fact that they are being recorded, but this is not always possible due to ethical considerations and resource constraints.

Due to ethical considerations and resource constraints, researchers may be forced to inform the human interlocutors that they are being recorded or to setup artificial experiments in which they hire humans and instruct them to carry out a particular task by interacting with a dialogue system. In these cases, there is no guarantee that the interactions in the corpus will reflect true interactions, since the hired humans may behave differently from the true user population. One factor that may cause behavioural differences is the fact that the hired humans may not share the same intentions and motivations as the true user population \citep{young2013pomdp}. 
The unnaturalness may be further exacerbated by the hiring process, as well as the platform through which they interact. Such factors are becoming more prevalent as researchers increasingly rely on crowdsourcing platforms, such as Amazon Mechanical Turk, to collect and evaluate dialogue data \citep{jurcicek2011real}. 

In the case of \textit{Wizard-of-Oz} experiments \citep{bohus2008sorry,petrik2004wizard}, a human thinks (s)he is speaking to a machine, but a human operator is in fact controlling the dialogue system.  This enables the generation of datasets that are closer in nature to the dialogues humans may wish to have with a good AI dialogue system. Unfortunately, such experiments are expensive and time-consuming to carry out. Ultimately the impact of any unnaturalness in the dialogues depends on the task and context in which the dialogue system is deployed. 


\subsection{Corpora from Fiction} \label{subsec:corpora_from_fiction}
It is also possible to use artificial dialogue corpora for data-driven learning.
This includes corpora based on works of fiction such as novels, movie manuscripts and audio subtitles.
However, unlike transcribed human-human conversations, novels, movie manuscripts, and audio subtitles depend upon events outside the current conversation, which are not observed.
This makes data-driven learning more difficult because the dialogue system has to account for unknown factors.
The same problem is also observed in certain other media, such as microblogging websites (e.g.\@ Twitter and Weibo), where conversations also may depend on external unobserved events.

Nevertheless, recent studies have found that spoken language in movies resembles spontaneous human spoken language \citep{forchini2009spontaneity}. Although movie dialogues are explicitly written to be spoken and contain certain artificial elements, many of the linguistic and paralinguistic features contained within the dialogues are similar to natural spoken language, including dialogue acts such as turn-taking and reciprocity (e.g.\@ returning a greeting when greeted). The artificial differences that exist may even be helpful for data-driven dialogue learning since movie dialogues are more compact, follow a steady rhythm, and contain less “garbling” and repetition, all while still presenting a clear event or message to the viewer \citep{dose2013flipping,forchini2009spontaneity,forchini2012movie}. Unlike dialogues extracted from \textit{Wizard-of-Oz} human experiments, movie dialogues span many different topics and occur in many different environments \citep{webb2010corpus}.
They contain different actors with different intentions and relationships to one another, which could potentially allow a data-driven dialogue system to learn to personalize itself to different users by making use of different interaction patterns \citep{li2016persona}.

\subsection{Corpus Size}
As in other machine learning applications such as machine translation \citep{al2000translating,DBLP:journals/corr/GulcehreFXCBLBS15} and speech recognition \citep{deng2013machine,bengio2014deep}, the size of the dialogue corpus is important for building an effective data-driven dialogue \citep{lowe2015ubuntu,2015arXiv150704808S}. 

There are two primary perspectives on the importance of dataset size for building data-driven dialogue systems. The first perspective comes from the machine learning literature: larger datasets place constraints on the dialogue model trained from that data. Datasets with few examples may require strong structural priors placed on the model, such as using a modular system, while large datasets can be used to train end-to-end dialogue systems with less \textit{a priori} structure.
The second comes from a statistical natural language processing perspective: since the statistical complexity of a corpus grows with the linguistic diversity and number of topics, the number of examples required by a machine learning algorithm to model the patterns in it will also grow with the linguistic diversity and number of topics.
Consider two small datasets with the same number of dialogues in the domain of bus schedule information: in one dataset the conversations between the users and operator is natural, and the operator can improvise and chitchat; in the other dataset, the operator reads from a script to provide the bus information.
Despite having the same size, the second dataset will have less linguistic diversity and not include chitchat topics.
Therefore, it will be easier to train a data-driven dialogue system mimcking the behaviour of the operator in the second dataset, however it will also exhibit a highly pedantic style and not be able to chitchat.
In addition to this, to have an effective discussion between any two agents, their common knowledge must be represented and understood by both parties. The process of establishing this common knowledge, also known as \textit{grounding}, is especially critical to repair misunderstandings between humans and dialogue systems \citep{cahn1999psychological}. Since the number of misunderstandings can grow with the lexical diversity and number of topics (e.g.\@ misunderstanding the paraphrase of an existing word, or misunderstanding a rarely seen keyword), the number of examples required to repair these grow with linguistic diversity and topics.
In particular, the effect of linguistic diversity has been observed in practice: \cite{vinyals2015neural} train a simple encoder-decoder neural network on a proprietary dataset of technical support dialogues. Although it has a similar size and purpose as the Ubuntu Dialogue Corpus~\citep{lowe2015ubuntu}, the qualitative examples shown by \cite{vinyals2015neural} are significantly superior to those obtained by more complex models on the Ubuntu Corpus \citep{serban2016multiresolution}.
This result may likely be explained in part due to the fact that technical support operators often follow a comprehensive script for solving problems. As such, the script would reduce the linguistic diversity of their responses.

Furthermore, since the majority of human-human dialogues are multi-modal and highly ambiguous in nature \citep{chartrand1999chameleon,de2013speaker}, the size of the corpus may compensate for some of the ambiguity and missing modalities. 
If the corpus is sufficiently large, then the resolved ambiguities and missing modalities may, for example, be approximated using latent stochastic variables~\citep{serban2016hierarchical}. Thus, we include corpus size as a dimension of analysis.
We also discuss the benefits and drawbacks of several popular large-scale datasets in Section \ref{sec:bigdata}.






\section{Available Dialogue Datasets}
There is a vast amount of data available documenting human communication. Much of this data could be used --- perhaps after some pre-processing --- to train a dialogue system. However, covering all such sources of data would be infeasible. Thus, we restrict the scope of this survey to datasets that have already been used to study dialogue or build dialogue systems, and to very large corpora of interactions---that may or may not be strictly considered dialogue datasets---which could be leveraged in the near future to build more sophisticated data-driven dialogue models. We restrict the selection further to contain only corpora generated from spoken or written English, and to corpora which, to the best of our knowledge, either are publicly available or will be made available in the near future. We first give a brief overview of each of the considered corpora, and later highlight some of the more promising examples, explaining how they could be used to further dialogue research.\footnote{We form a live list of the corpora discussed in this work, along with links to downloads, at: \url{http://breakend.github.io/DialogDatasets}. Pull requests can be made to the Github repository (\url{https://github.com/Breakend/DialogDatasets}) hosting the website for continuing updates to the list of corpora.}

The dialogue datasets analyzed in this paper are listed in Tables~\ref{tab:hc}-\ref{tab:hh_written}. 
Column features indicate properties of the datasets, including the number of dialogues, average dialogue length, number of words, whether the interactions are between humans or with an automated system, and whether the dialogues are written or spoken. Below, we discuss qualitative features of the datasets, while statistics can be found in the aforementioned table.

\subsection{Human-Machine Corpora}
As discussed in Subsection \ref{subsec:humanhuman_vs_humansystem_corpora}, an important distinction between dialogue datasets is whether they consist of dialogues between two humans or between a human and a machine. Thus, we begin by outlining some of the existing human-machine corpora in several categories based on the types of systems the humans interact with: Restaurant and Travel Information, Open-Domain Knowledge Retrieval, and Other Specialized systems. Note, we also include human-human corpora here where one human plays the role of the machine in a Wizard-of-Oz fashion.

\subsubsection{Restaurant and Travel Information}

One common theme in human-machine language datasets is interaction with systems which provide restaurant or travel information. Here we'll briefly describe some human-machine dialogue datasets in this domain.

One of the most popular recent sources of such data has come from the datasets for structured dialogue prediction released in conjunction with the\textbf{ Dialog State Tracking Challenge} (DSTC) \citep{williams2013dialog}. As the name implies, these datasets are used to learn a strategy for the Dialogue State Tracker (sometimes called `belief tracking'), which involves estimating the intentions of a user throughout a dialog. State tracking is useful as it can increase the robustness of speech recognition systems, and can provide an implementable framework for real-world dialogue systems. Particularly in the context of goal-oriented dialogue systems (such as those providing travel and restaurant information), state tracking is necessary for creating coherent conversational interfaces. As such, the first three datasets in the DSTC---referred to as \textbf{DSTC1, DSTC2, and DSTC3} respectively---are medium-sized spoken datasets obtained from human-machine interactions with restaurant and travel information systems. All datasets provide labels specifying the current goal and desired action of the system.

DSTC1 \citep{williams2013dialog} features conversations with an automated bus information interface, where users request bus routes from the system and the system responds with clarifying queries or the desired information. DSTC2 introduces changing user goals in a restaurant booking system, while trying to provide a desired reservation\citep{henderson2014second}. DSTC3 introduces a small amount of labelled data in the domain of tourist information. It is intended to be used in conjunction with the DSTC2 dataset as a domain adaptation problem \citep{henderson2014dialog}.


The \textbf{Carnegie Mellon Communicator Corpus} \citep{bennett2002carnegie} also contains human-machine interactions with a travel booking system. It is a medium-sized dataset of interactions with a system providing up-to-the-minute flight information, hotel information, and car rentals. Conversations with the system were transcribed, along with the user's comments at the end of the interaction.

The \textbf{ATIS (Air Travel Information System) Pilot Corpus} \citep{hemphill1990atis} is one of the first human-machine corpora. It consists of interactions, lasting about 40 minutes each, between human participants and a travel-type booking system, secretly operated by humans. Unlike the \textbf{Carnegie Mellon Communicator Corpus}, it only contains $1041$ utterances.

In the \textbf{Maluuba Frames Corpus} \citep{elframes}, one user plays the role of a conversational agent in a Wizard-of-Oz fashion, while the other user is tasked with finding available travel or vacation accommodations according to a pre-specified task. The Wizard is provided with a knowledge database which recorded their actions. Semantic frames are annotated in addition to actions which the Wizard performed on the database to accompany a line of dialogue. In this way, the Frames corpus aims to track decision-making processes in travel- and hotel-booking through natural dialog.


\subsubsection{Open-Domain Knowledge Retrieval}

Knowledge retrieval and Question \& Answer (QA) corpora are a broad distinction of corpora that we will not extensively review here. Instead, we include only those QA corpora which explicitly record interactions of humans with existing systems. The \textbf{Ritel corpus} \citep{rosset2006ritel} is a small dataset of 528 dialogs with the Wizard-of-Oz Ritel platform. The project's purpose was to integrate spoken language dialogue systems with open-domain information retrieval systems, with the end goal of allowing humans to ask general questions and iteratively refine their search. The questions in the corpus mostly revolve around politics and the economy, such as ``Who is currently presiding the Senate?'', along with some conversations about arts and science-related topics.

Other similar open-domain corpora in this area include WikiQA \cite{yang2015wikiqa} and MS MARCO \cite{nguyen2016ms}, which compile responses from automated Bing searches and human annotators. However, these do not record dialogs, but rather simply gather possible responses to queries. As such, we won't discuss these datasets further, but rather mention them briefly as examples of other Open-Domain corpora in the field.

\begin{landscape}
\begin{table}
\centering
\footnotesize
\begin{tabular}{|l|l|l|c|c|c|l|} \hline
Name & Type & Topics & Avg. \# & Total \# & Total \# & Description \\
& & & of turns & of dialogues & of words & \\
\hline \hline
DSTC1 \citep{williams2013dialog} & Spoken & Bus schedules & 13.56 & 15,000 & 3.7M & Bus ride information system \\ \hline
DSTC2 \citep{henderson2014second} & Spoken & Restaurants & 7.88 & 3,000 & 432K & Restaurant booking system \\ \hline
DSTC3 \citep{henderson2014dialog} & Spoken & Tourist information& 8.27 & 2,265 & 403K & Information for tourists \\ \hline
CMU Communicator Corpus & Spoken & Travel & 11.67 & 15,481 & 2M* & Travel planning and booking system \\
\citep{bennett2002carnegie} &&&&&& \\ \hline
ATIS Pilot Corpus\textsuperscript{\textdagger} & Spoken & Travel & 25.4 & 41 & 11.4K* & Travel planning and booking system \\
\citep{hemphill1990atis} &&&&&& \\ \hline
Ritel Corpus\textsuperscript{\textdagger} & Spoken & Unrestricted/ Diverse Topics & 9.3* & 582 & 60k & An annotated open-domain question \\
\citep{rosset2006ritel}&&&&&& answering spoken dialogue system \\ \hline
DIALOG Mathematical & Spoken & Mathematics & 12 & 66 & 8.7K* & Humans interact with computer system \\
Proofs \citep{wolska2004annotated} &&&&&& to do mathematical theorem proving \\ \hline
MATCH Corpus\textsuperscript{\textdagger} & Spoken & Appointment Scheduling & 14.0 & 447 & 69K* & A system for scheduling appointments. \\
\citep{georgila2010match}&&&&&& Includes dialogue act annotations \\ \hline
Maluuba Frames\textsuperscript{\textdagger} & Chat, QA \& & Travel \& Vacation & 15 & 1369 & -- & For goal-driven dialogue systems.\\
\citep{elframes} & Recommendation & Booking & & & &Semantic frames labeled and actions\\
& & & & & &taken on a knowledge-base annotated.\\\hline

\end{tabular}
\caption{Human-machine dialogue datasets. Starred (*) numbers are approximated based on the average number of words per utterance. Datasets marked with (\textsuperscript{\textdagger}) indicate Wizard-of-Oz dialogues, where the machine is secretly operated by a human.}
\label{tab:hc}
\end{table}

\begin{table}
\centering
\footnotesize
\begin{tabular}{|l|l|c|c|c|l|} \hline
Name &  Topics & Total \# & Total \# & Total& Description \\
&  & of dialogues & of words & length & \\
\hline \hline
HCRC Map Task Corpus &	Map-Reproducing & 128 & 147k & 18hrs &Dialogues from HLAP Task in which speakers must collaborate verbally \\
\citep{anderson1991hcrc}& Task&&&& to reproduce on one participant’s map a
route printed on the other’s. \\ \hline
The Walking Around Corpus &  	Location &	36 & 300k* & 33hrs & People collaborating over telephone to find certain locations. \\
\citep{brennan2013entrainment}& Finding Task & & & & \\ \hline
Green Persuasive Database &	  	Lifestyle & 8 & 35k* & 4hrs & A persuader with (genuinely) strong pro-green feelings tries to
 convince \\
\citep{douglas2007humaine} &&&&& persuadees to consider adopting more ‘green’ lifestyles. \\ \hline
Intelligence Squared Debates &	  	Debates & 108 & 1.8M & 200hrs* & Various topics in Oxford-style debates, each constrained\\
\citep{zhang2016conversational} &&&&& to one subject. Audience opinions provided pre- and post-debates.  \\ \hline
The Corpus of Professional Spoken&   	Politics, Education & 200 & 2M & 220hrs* & Interactions from faculty meetings and White House \\
American English \citep{barlow2000corpus}&&&&&  press conferences.\\ \hline
MAHNOB Mimicry Database &   	Politics, Games & 54 & 100k* & 11hrs &	Two experiments: a discussion on a political topic, and a \\
\citep{sun2011multimodal} &&&&&role-playing game.\\ \hline
The IDIAP Wolf Corpus &   	Role-Playing Game & 15 & 60k* & 7hrs & A recording of Werewolf role-playing game with annotations \\
\citep{hung2010idiap}&&&&&related to game progress.\\ \hline
SEMAINE corpus &    	Emotional  & 100 & 450k* & 50hrs &	Users were recorded while holding conversations with an operator  \\
\citep{mckeown2010semaine}& Conversations &&&& who
adopts roles designed to evoke emotional reactions.\\ \hline
DSTC4/DSTC5 Corpora&    	Tourist	 & 35 & 273k & 21hrs & Tourist information exchange over Skype. \\
\citep{kim2015dialog,kim2016fifth}&&&&&\\\hline
Loqui Dialogue Corpus  & Library Inquiries & 82 & 21K & 140* & Telephone interactions between librarians and patrons.\\
\citep{passonneau2014loqui} &&&&& Annotated dialogue acts, discussion topics, frames (discourse units),  \\
&&&&&question-answer pairs.  \\ \hline
MRDA Corpus& ICSI Meetings & 75 & 11K* & 72hrs & Recordings of ICSI meetings. Topics include: the corpus project itself,  \\
\citep{shriberg2004icsi}&&&&&automatic speech recognition, natural language
processing and theories of\\
&&&&&language. Dialogue acts, question-answer pairs, and hot spots. \\ \hline
TRAINS 93 Dialogues Corpus& Railroad Freight&98 & 55K & 6.5hrs & Collaborative planning of railroad freight routes.\\
\citep{heeman1995trains}&Route Planning&&&&\\\hline
Verbmobil Corpus  &  Appointment & 726  & 270K & 38Hrs & Spontaneous speech data collected for the Verbmobil project.\\   
\citep{weilhammer2002multi} &Scheduling&&&&Full corpus is in English, German, and Japanese. \\
&&&&& We only show English statistics. \\ \hline
\end{tabular}
\caption{Human-human constrained spoken dialogue datasets. Starred (*) numbers are estimates based on the average rate of English speech from the National Center for Voice and Speech (\small{\url{www.ncvs.org/ncvs/tutorials/voiceprod/tutorial/quality.html}}).}
\label{tab:hh_constrained_spoken}
\end{table}

\end{landscape}
\begin{landscape}

\begin{table}
\centering
\footnotesize
\begin{tabular}{|l|l|c|c|c|l|} \hline
Name &  Topics & Total \# & Total \# & Total & Description \\
&  & of dialogues & of words & length & \\
\hline \hline
Switchboard	\citep{godfrey1992switchboard}& 	Casual Topics & 2,400 & 3M & 300hrs* &	Telephone conversations on pre-specified topics	\\ \hline
British National Corpus (BNC) &	  	Casual Topics	 & 854 & 10M & 1,000hrs* & British dialogues many contexts, from formal business or government\\
\citep{leech1992100}&&&&& meetings to radio shows and phone-ins. \\ \hline
CALLHOME American English &	Casual Topics & 120 & 540k* & 60hrs & Telephone conversations between family members or close friends.\\
Speech \citep{canavan1997callhome}&&&&& \\ \hline
CALLFRIEND American English  &		Casual Topics & 60 & 180k* & 20hrs & Telephone conversations between Americans with a Southern accent. \\
Non-Southern Dialect &&&&& \\
\citep{canavan1996callfriend} &&&&& \\ \hline
The Bergen Corpus of London  &	 Unrestricted & 100 & 500k & 55hrs &Spontaneous teenage talk recorded in 1993. \\
Teenage Language &&&&& Conversations were recorded secretly.\\
\citep{haslerud1995bergen} &&&&& \\ \hline
The Cambridge and Nottingham
 & 	Casual Topics & -- & 5M & 550hrs* & British dialogues from wide variety of informal contexts, such as \\
 Corpus of Discourse in English &&&&&hair salons, restaurants, etc. \\
 \citep{mccarthy1998spoken} &&&&& \\ \hline
D64 Multimodal Conversation Corpus & 	Unrestricted & 2 & 70k* & 8hrs & Several hours of natural interaction between a group of people \\
\citep{oertel2013d64} &&&&& \\ \hline
AMI Meeting Corpus &  Meetings & 175 & 900k* & 100hrs &	Face-to-face meeting
recordings.	\\
\citep{renals2007recognition} &&&&& \\ \hline
Cardiff Conversation Database & Unrestricted & 30 & 20k* & 150min &Audio-visual database with unscripted natural conversations, \\
(CCDb) \citep{aubrey2013cardiff} &&&&&including visual annotations. \\ \hline
4D Cardiff Conversation Database & Unrestricted &17 & 2.5k*& 17min& A version of the CCDb with 3D video\\
(4D CCDb) \citep{vandeventer20154d} &&&&& \\ \hline
The Diachronic Corpus of  &  Casual Topics & 280 & 800k & 80hrs* & Selection of face-to-face, telephone, and public\\
Present-Day Spoken English &&&&&discussion dialogue from Britain. \\
\citep{aarts2006diachronic} &&&&& \\ \hline
The Spoken Corpus of the  &		Casual Topics & 314 & 800k & 60hrs &Dialogue of people aged 60 or above talking about their memories, \\
Survey of English Dialects &&&&& families, work and the folklore of the countryside from a century ago.\\
\citep{beare1999spoken} &&&&& \\ \hline
The Child Language Data &	Unrestricted & 11K & 10M & 1,000hrs* &International database organized for the  \\
Exchange System  &&&&& study of first and second language acquisition. \\
\citep{macwhinney1985child} &&&&& \\ \hline
The Charlotte Narrative and  &		Casual Topics & 95 & 20K & 2hrs* &Narratives, conversations and interviews representative  \\
Conversation Collection (CNCC) &&&&& of the residents of Mecklenburg County, North Carolina. \\
\citep{reppen2004american} &&&&& \\ \hline

\end{tabular}
\caption{Human-human spontaneous spoken dialogue datasets. Starred (*) numbers are estimates based on the average rate of English speech from the National Center for Voice and Speech (\small{\url{www.ncvs.org/ncvs/tutorials/voiceprod/tutorial/quality.html}})}
\label{tab:hh_spontaneous_spoken}
\end{table}

\end{landscape}
\begin{landscape}

\begin{table}
\centering
\footnotesize
\begin{tabular}{|l|l|c|c|c|c|l|} \hline
Name & Topics & Total \# & Total \# & Total \# & Total \# & Description \\
&  & of utterances & of dialogues & of works & of words & \\
\hline \hline
Movie-DiC &	 	Movie	 &		764k & 	132K	& 753 & 6M & Movie scripts of American films.\\
\citep{banchs2012movie} &dialogues&&&&& \\ \hline
Movie-Triples &	 	Movie &	736k &	245K	& 614 & 13M & Triples of utterances which are filtered to come\\
\citep{2015arXiv150704808S} & dialogues&&&&&from X-Y-X triples. \\ \hline
Film Scripts Online Series & Movie & 1M* & 263K\textsuperscript{\textdagger} & 1,500 & 16M* & Two subsets of scripts\\
&scripts &&&&& (1000 American films and 500 mixed British/American films). \\ \hline
Cornell Movie-Dialogue Corpus &   Movie & 305K & 220K & 617 & 9M* & Short conversations from film scripts, annotated\\
\citep{Danescu-Niculescu-Mizil+Lee:11a}&dialogues&&&&& with character metadata. \\ \hline
Filtered Movie Script Corpus &   Movie & 173k & 87K & 1,786 & 2M* & Triples of utterances which are filtered to come\\
\citep{nio2014conversation} &dialogues&&&&&from X-Y-X triples. \\ \hline
American Soap Opera  & TV show & 10M* & 1.2M\textsuperscript{\textdagger} & 22,000 & 100M & Transcripts of American soap operas.\\
Corpus \citep{davies2012corpus}&scripts&&&&& \\ \hline
TVD Corpus&  TV show	& 60k* &	10K\textsuperscript{\textdagger}	& 191 & 600k* & TV scripts from a comedy (Big Bang Theory) and\\
\citep{roy2014tvd} & scripts&&&&& drama (Game of Thrones) show.\\ \hline
Character Style from Film&  Movie	& 664k & 151K & 862 & 9.6M & Scripts from IMSDb, annotated for linguistic\\
Corpus \citep{walker2012annotated} &scripts&&&&& structures and character archetypes.\\ \hline
SubTle Corpus&  Movie	& 6.7M &	3.35M	&6,184 & 20M & Aligned interaction-response pairs from \\
\citep{ameixa2013subtitles} &subtitles&&&&&movie subtitles. \\ \hline
OpenSubtitles &   Movie & 140M* &	36M\textsuperscript{\textdagger}	 & 207,907 & 1B & Movie subtitles which are not speaker-aligned.
\\
 \citep{tiedemann2012parallel} &subtitles&&&&& \\ \hline
CED (1560--1760) Corpus & Written Works & -- & -- & 177 & 1.2M & Various scripted fictional works from (1560--1760)
\\
 \citep{kyto2006guide} &\& Trial Proceedings&&&&& as well as court trial proceedings. \\ \hline
\end{tabular}
\caption{Human-human scripted dialogue datasets. Quantities denoted with (\textsuperscript{\textdagger}) indicate estimates based on average number of dialogues per movie \citep{banchs2012movie} and the number of scripts or works in the corpus. Dialogues may not be explicitly separated in these datasets. TV show datasets were adjusted based on the ratio of average film runtime (112 minutes) to average TV show runtime (36 minutes). This data was scraped from the IMBD database (\small{\url{http://www.imdb.com/interfaces}}). ( Starred (*) quantities are estimated based on the average number of words and utterances per film, and the average lengths of films and TV shows. Estimates derived from the Tameri Guide for Writers (\small{\url{http://www.tameri.com/format/wordcounts.html}}).}
\label{tab:hh_scripted}
\end{table}
\end{landscape}
\begin{landscape}

\begin{table}
\centering
\footnotesize
\begin{tabular}{|l|l|l|c|c|c|l|} \hline
Name & Type & Topics & Avg. \# & Total \# & Total \# & Description \\
& & & of turns & of dialogues & of words & \\
\hline \hline
NPS Chat Corpus & Chat & Unrestricted & ~704 & 15\textsuperscript{\textdagger} & 100M & Posts from age-specific online chat rooms. \\
\citep{forsyth2007lexical} &&&&&& \\ \hline
Twitter Corpus & Microblog & Unrestricted & 2 & 1.3M & ~125M\textsuperscript{\textdaggerdbl} & Tweets and replies extracted from Twitter \\
\citep{Ritter:2010:UMT:1857999.1858019} &&&&&& \\ \hline
Twitter Triple Corpus & Microblog & Unrestricted & 3 & 4,232 & ~65K\textsuperscript{\textdaggerdbl} & A-B-A triples extracted from Twitter \\
\citep{sordoni2015aneural} &&&&&& \\ \hline
UseNet Corpus & Microblog & Unrestricted & ~687 & 47860\textsuperscript{\textdagger} & ~7B & UseNet forum postings \\
\citep{shaoul2009usenet} &&&&&& \\ \hline
NUS SMS Corpus  & SMS messages & Unrestricted & ~18 & ~3K & 580,668*{$^\Box$} & SMS messages collected between two  \\
\citep{chen2013creating}&&&&&& users, with timing analysis. \\ \hline
Reddit\degree & Forum & Unrestricted & -- & -- & -- & 1.7B comments across Reddit. \\\hline
Reddit Domestic Abuse Corpus & Forum & Abuse help & 17.53 & 21,133 & 19M-103M {$^\triangle$} & Reddit posts from either domestic abuse \\
\citep{ray4183analysis} &&&&&&subreddits, or general chat. \\ \hline
Settlers of Catan & Chat & Game terms & ~95 & 21 & -- & Conversations between players \\
\citep{afantenos2012developing}&&&&&& in the game `Settlers of Catan.'\\ \hline
Cards Corpus & Chat & Game terms & 38.1 & 1,266 & 282K & Conversations between players \\
\citep{djalali2012corpus} &    &   & & & & playing `Cards world.' \\ \hline
Agreement in Wikipedia Talk Pages & Forum & Unrestricted & 2 & 822 & 110K & LiveJournal and Wikipedia Discussions forum threads. \\
\citep{andreas2012annotating}& & & & & &Agreement type and level annotated.\\ \hline
Agreement by Create Debaters & Forum & Unrestricted &  2& 10K & 1.4M & Create Debate forum conversations. Annotated what type\\
\citep{rosenthal2015couldn}& & & & & &of agreement (e.g. paraphrase) or disagreement.\\ \hline
Internet Argument Corpus & Forum & Politics & ~35.45 & ~11K & ~73M & Debates about specific \\ \citep{walker2012corpus}&&&&&& political or moral positions. \\ \hline
MPC Corpus & Chat & Social tasks & 520 & 14 & 58K & Conversations about general, \\
\citep{shaikh2010mpc}&&&&&& political, and interview topics. \\ \hline
Ubuntu Dialogue Corpus & Chat & Ubuntu Operating & 7.71 & 930K & 100M & Dialogues extracted from \\
\citep{lowe2015ubuntu} &&System&&&&Ubuntu chat stream on IRC. \\ \hline
Ubuntu Chat Corpus & Chat & Ubuntu Operating & ~3381.6 & 10665\textsuperscript{\textdagger} & ~2B*{$^\Box$} & Chat stream scraped from \\
\citep{uthus2013ubuntu}&&System&&&& IRC logs (no dialogues extracted). \\ \hline
Movie Dialog Dataset & Chat, QA \& & Movies & ~3.3 & ~3.1M{$^\blacktriangledown$} & ~185M & For goal-driven dialogue systems. Includes \\ \citep{dodge2015evaluating} & Recommendation & & & & & movie metadata as knowledge triples. \\ \hline
\end{tabular}
\caption{\small{Human-human written dialogue datasets. Starred (*) quantities are computed using word counts based on spaces (i.e.\@ a word must be a sequence of characters preceded and followed by a space), but for certain corpora, such as IRC and SMS corpora, proper English words are sometimes concatenated together due to slang usage. Triangle ({$^\triangle$}) indicates lower and upper bounds computed using average words per utterance estimated on a similar Reddit corpus \citet{schrading2015analyzing}. Square ({$^\Box$}) indicates estimates based only on the English part of the corpus. Note that 2.1M dialogues from the Movie Dialog dataset ({$^\blacktriangledown$}) are in the form of simulated QA pairs. Dialogs indicated by (\textsuperscript{\textdagger}) are contiguous blocks of recorded conversation in a multi-participant chat. In the case of UseNet, we note the total number of newsgroups and calculate the average turns as the average number of posts collected per newsgroup. (\textsuperscript{\textdaggerdbl}) indicates an estimate based on a Twitter dataset of similar size and refers to tokens as well as words. (\degree) refers to: \footnotesize{\url{https://www.reddit.com/r/datasets/comments/3bxlg7/i\_have\_every\_publicly\_available\_reddit\_comment/}}}}
\label{tab:hh_written}
\end{table}

\end{landscape}

\subsubsection{Other}

The \textbf{DIALOG mathematical proof dataset} \citep{wolska2004annotated} is a \textit{Wizard-of-Oz} dataset involving an automated tutoring system that attempts to advise students on proving mathematical theorems. This is done using a hinting algorithm that provides clues when students come up with an incorrect answer. At only 66 dialogues, the dataset is very small, and consists of a conglomeration of text-based interactions with the system, as well as think-aloud audio and video footage recorded by the users as they interacted with the system. The latter was transcribed and annotated with simple speech acts such as `signaling emotions' or `self-addressing'.

The \textbf{MATCH corpus} \citep{georgila2010match} is a small corpus of 447 dialogues based on a \textit{Wizard-of-Oz} experiment, which collected 50 young and old adults interacting with spoken dialogue systems. These conversations were annotated semi-automatically with dialogue acts and ``Information State Update'' (ISU) representations of dialogue context. The corpus also contains information about the users' cognitive abilities, with the motivation of modeling how the elderly interact with dialogue systems.

\subsection{Human-Human Spoken Corpora}

Naturally, there is much more data available for conversations between humans than conversations between humans and machines. Thus, we break down this category further, into spoken dialogues (this section) and written dialogues (Section \ref{sec:hhwrit}). The distinction between spoken and written dialogues is important, since the distribution of utterances changes dramatically according to the nature of the interaction. As discussed in Subsection \ref{subsec:spoken_versus_written}, spoken dialogues tend to be more colloquial and generally well-formed as the user speaks in train-of-thought manner; they also tend to use shorter words and phrases. Conversely, in written communication, users have the ability to reflect on what they are writing before they send a message. Written dialogues can also contain spelling errors or abbreviations, though, which are generally not transcribed in spoken dialogues.

\subsubsection{Spontaneous Spoken Corpora}

We first introduce datasets in which the topics of conversation are either casual, or not pre-specified in any way. We refer to these corpora as \textit{spontaneous}, as we believe they most closely mimic spontaneous and unplanned spoken interactions between humans.

Perhaps one of the most influential spoken corpora is the \textbf{Switchboard dataset} \citep{godfrey1992switchboard}. This dataset consists of approximately 2,500 dialogues from phone calls, along with word-by-word transcriptions with about 500 total speakers. A computer-driven robot operator system introduced a topic for discussion between two participants, and recorded the resulting conversation. About 70 casual topics were provided, of which about 50 were frequently used. The corpus was originally designed for training and testing various speech processing algorithms; however, it has since been used for a wide variety of other tasks, including the modeling of dialogue acts such as `statement', `question', and `agreement' \citep{stolcke2000dialogue}.

Another important dataset is the \textbf{British National Corpus} (BNC) \citep{leech1992100}, which contains approximately 10 million words of dialogue. These were collected in a variety of contexts ranging from formal business or government meetings, to radio shows and phone-ins. Although most of the conversations are spoken in nature, some of them are also written. BNC covers a large number of sources, and was designed to represent a wide cross-section of British English from the late twentieth century. The corpus also includes part-of-speech (POS) tagging for every word. The vast array of settings and topics covered by this corpus renders it very useful as a general-purpose spoken dialogue dataset.

Other datasets have been collected for the analysis of spoken English over the telephone. The \textbf{CALLHOME American English Speech Corpus} \citep{canavan1997callhome} consists of 120 such conversations totalling about 60 hours, mostly between family members or close friends. Similarly, the \textbf{CALLFRIEND American English-Non-Southern Dialect Corpus} \citep{canavan1996callfriend} consists of 60 telephone conversations lasting 5-30 minutes each between English speakers in North America without a Southern accent. It is annotated with speaker information such as sex, age, and education. The goal of the project was to support the development of language identification technologies, yet, there are no distinguishing features in either of these corpora in terms of the topics of conversation.

An attempt to capture exclusively teenage spoken language was made in the \textbf{Bergen Corpus of London Teenager Language} (COLT) \citep{haslerud1995bergen}. Conversations were recorded surreptitiously by student `recruits', with a Sony Walkman and a lapel microphone, in order to obtain a better representation of teenager interactions `in-the-wild'. This dataset has been used to identify trends in language evolution in teenagers \citep{stenstrom2002trends}.


The \textbf{Cambridge and Nottingham Corpus of Discourse in English} (CANCODE) \citep{mccarthy1998spoken} is a subset of the Cambridge International Corpus, containing about 5 million words collected from recordings made throughout the islands of Britain and Ireland. It was constructed by Cambridge University Press and the University of Nottingham using dialogue data on general topics between 1995 and 2000. It focuses on interpersonal communication in a
range of social contexts, varying from hair salons, to post offices, to restaurants. This has been used, for example, to study language awareness in relation
to spoken texts and their cultural contexts \citep{carter1998orders}. In the dataset, the relationships between speakers (e.g. roommates, strangers) is labeled and the interaction type is provided (e.g. professional, intimate).

Other works have attempted to record the physical elements of conversations between humans. To this end, a small corpus entitled \textbf{d64 Multimodal Conversational Corpus} \citep{oertel2013d64} was collected, incorporating data from 7 video cameras, and the registration of 3-D head, torso, and arm motion using an Optitrack system. Significant effort was made to make the data collection process as non-intrusive---and thus, naturalistic---as possible. Annotations were made to attempt to quantify overall group excitement and pairwise social distance between participants.

A similar attempt to incorporate computer vision features was made in the \textbf{AMI Meeting Corpus} \citep{renals2007recognition}, where cameras, a VGA data projector capture, whiteboard capture, and digital pen capture, were all used in addition to speech recordings for various meeting scenarios. As with the d64 corpus, the AMI Meeting Corpus is a small dataset of multi-participant chats, that has not been disentangled into strict dialogue. The dataset has often been used for analysis of the dynamics of various corporate and academic meeting scenarios.

In a similar vein, the \textbf{Cardiff Conversation Database} (CCDb) \citep{aubrey2013cardiff} is an audio-visual database containing unscripted natural conversations between pairs of people. The original dataset consisted of 30 five minute conversations, 7 of which were fully annotated with transcriptions and behavioural annotations such as speaker activity, facial expressions, head motions, and smiles. The content of the conversation is an unconstrained discussion on topics such as movies. While the original dataset featured 2D visual feeds, an updated version with 3D video has also been derived, called the \textbf{4D Cardiff Conversation Database} (4D CCDb) \citep{vandeventer20154d}. This version contains 17 one-minute conversations from 4 participants on similarly un-constrained topics.

The \textbf{Diachronic Corpus of Present-Day Spoken English} (DCPSE) \citep{aarts2006diachronic} is a parsed corpus of spoken English made up of two separate datasets. It contains more than 400,000 words from the ICE-GB corpus (collected in the early 1990s) and 400,000 words from the London-Lund Corpus (collected in the late
1960s-early 1980s). ICE-GB refers to the British component of the International Corpus of English \citep{greenbaum1996international, greenbaum1996comparing} and contains both spoken and written dialogues from English adults who have completed secondary education. The dataset was selected to provide a representative sample of British English. The London-Lund Corpus \citep{svartvik1990london} consists exclusively of spoken British conversations, both dialogues and monologues. It contains a selection of face-to-face, telephone, and public discussion dialogues; the latter refers to dialogues that are heard by an audience that does not participate in the dialogue, including interviews and panel discussions that have been broadcast.  The orthographic transcriptions of the datasets are normalised and annotated according to the same criteria; ICE-GB was used as a gold standard for the parsing of DCPSE.

The \textbf{Spoken Corpus of the Survey of English Dialects} \citep{beare1999spoken} consists of 1000 recordings, with about 0.8 million total words, collected from 1948-1961 in order to document various existing English dialects. People aged 60 and over were recruited, being most likely to speak the traditional `uncontaminated' dialects of their area and encouraged to talk about their memories, families, work, and their countryside folklore.

The \textbf{Child Language Data Exchange System} (CHILDES) \citep{macwhinney1985child}  is a database
organized for the study of first and second language acquisition. The database contains 10 million English words and approximately the same number of non-English words. It also contains transcripts, with occasional audio and video recordings of data collected from children and adults learning both first and second languages, although the English transcripts are mostly from children. This corpus could be leveraged in order to build automated teaching assistants.

The expanded \textbf{Charlotte Narrative and Conversation Collection} (CNCC), a subset of the first release of the American National Corpus \citep{reppen2004american}, contains 95 narratives, conversations and interviews representative of the residents of Mecklenburg County, North Carolina and its surrounding communities. The purpose of the CNCC was to create a corpus of conversation and conversational narration in a 'New South' city at the beginning of the 21st century, that could be used as a resource for linguistic analysis. It was originally released as one of several collections in the New South Voices corpus, which otherwise contained mostly oral histories.  Information on speaker age and gender in the CNCC is included in the header for each transcript.

\subsubsection{Constrained Spoken Corpora}\label{sec:conspok}

Next, we discuss domains in which conversations only occur about a particular topic, or intend to solve a specific task. Not only is the topic of the conversation specified beforehand, but participants are discouraged from deviating off-topic. As a result, these corpora are slightly less general than their spontaneous counterparts; however, they may be useful for building goal-oriented dialogue systems. As discussed in Subsection \ref{subsec:natural_vs_unnatural_corpora}, this may also make the conversations less natural. We can further subdivide this category into the types of topics they cover: path-finding or planning tasks, persuasion tasks or debates, Q\&A or information retrieval tasks, and miscellaneous topics.

\paragraph{Collaborative Path-Finding or Planning Tasks} Several corpora focus on task planning or path-finding through the collaboration of two interlocutors. In these corpora typically one person acts as the decision maker and the other acts as the observer.

A well-known example of such a dataset is the \textbf{HCRC Map Task Corpus} \citep{anderson1991hcrc}, that consists of unscripted, task-oriented dialogues that have been digitally recorded and transcribed. The corpus uses the Map Task \citep{brown1984teaching}, where participants must collaborate verbally to reproduce a route on one of the participant's map on the map of another participant. The corpus is fairly small, but it controls for the familiarity between speakers, eye contact between speakers, matching between landmarks on the participants' maps, opportunities for contrastive stress, and phonological characteristics of landmark names. By adding these controls, the dataset attempts to focus on solely the dialogue and human speech involved in the planning process. 

The \textbf{Walking Around Corpus} \citep{brennan2013entrainment} consists of 36 dialogues between people communicating over mobile telephone. The dialogues have two parts: first, a `stationary partner' is asked to direct a `mobile partner' to find 18 destinations on a medium-sized university campus. The stationary partner is equipped with a map marked with the target destinations accompanied by photos of the locations, while the mobile partner is given a GPS navigation system and a camera to take photos. In the second part, the participants are asked to interact in-person in order to duplicate the photos taken by the mobile partner. The goal of the dataset is to provide a testbed for natural lexical entrainment, and to be used as a resource for pedestrian navigation applications.

The \textbf{TRAINS 93 Dialogues Corpus} \citep{heeman1995trains} consists of recordings of two interlocutors interacting to solve various planning tasks for scheduling train routes and arranging railroad freight. One user acts the role of a planning assistant system and the other user acts as the coordinator. This was not done in a Wizard-of-Oz fashion, and as such is not considered a Human-Machine corpus. 34 different interlocutors were asked to complete 20 different tasks such as: ``Determine the maximum number of boxcars of oranges that you could get to Bath by 7 AM tomorrow morning. It is now 12 midnight.'' The person playing the role of the planning assistant was provided with access to information that is needed to solve the task. Also included in the dataset is the information available to both users, the length of dialogue, and the speaker and ``system'' interlocutor identities.

The \textbf{Verbmobil Corpus} \citep{weilhammer2002multi} is a multilingual corpus consisting of English, German, and Japanese dialogues collected for the purposes of training and testing the Verbmobil project system. The system was a designed for speech-to-speech machine translation tasks. Dialogues were recorded in a variety of conditions and settings with room microphones, telephones, or close microphones, and were subsequently transcribed. Users were tasked with planning and scheduling an appointment throughout the course of the dialogue. Note that while there have been several versions of the Verbmobil corpora released, we refer to the entire collection here as described in \citep{weilhammer2002multi}. Dialogue acts were annotated in a subset of the corpus (1,505 mixed dialogues in German, English and Japanese). 76,210 acts were annotated with 32 possible categories of dialogue acts \cite{alexandersson2000modeling}\footnote{Note, this information and further facts about the Verbmobil project and corpus can be found here: \url{http://verbmobil.dfki.de/facts.html}}.

\paragraph{Persuasion and Debates} Another theme recurring among constrained spoken corpora is the appearance of persuasion or debate tasks. These can involve general debates on a topic or tasking a specific interlocutor to try to convince another interlocutor of some opinion or topic. Generally, these datasets record the outcome of how convinced the audience is of the argument at the end of the dialogue or debate.

The \textbf{Green Persuasive Dataset} \citep{douglas2007humaine} was recorded in 2007 to provide data for the HUMAINE project, whose goal is to develop interfaces that can register and respond to emotion. In the dataset, a persuader with strong pro-environmental (`pro-green') feelings tries to convince persuadees to consider adopting more ‘green’ lifestyles; these interactions are in the form of dialogues. It contains 8 long dialogues, totalling about 30 minutes each. Since the persuadees often either disagree or agree strongly with the persuader’s points, this would be good corpus for studying social signs of (dis)-agreement between two people.

The \textbf{MAHNOB Mimicry Database} \citep{sun2011multimodal} contains 11 hours of recordings, split over 54 sessions between 60 people engaged either in a socio-political discussion or negotiating a tenancy agreement. This dataset consists of a set of fully synchronised audio-visual recordings of natural dyadic (one-on-one) interactions. It is one of several dialogue corpora that provide multi-modal data for analyzing human behaviour during conversations. Such corpora often consist of auditory, visual, and written transcriptions of the dialogues. Here, only audio-visual recordings are provided. The purpose of the dataset was to analyze mimicry  (i.e.\@ when one participant mimics the verbal and nonverbal expressions of their counterpart). The authors provide some benchmark video classification models to this effect.

The \textbf{Intelligence Squared Debate Dataset} \citep{zhang2016conversational} covers the ``Intelligence Squared'' Oxford-style debates taking place between 2006 and 2015. The topics of the debates vary across the dataset, but are constrained within the context of each debate.
Speakers are labeled and the full transcript of the debate is provided. Furthermore, the outcome of the debate is provided (how many of the audience members were for the given proposal or against, before and after the debate).

\paragraph{QA or Information Retrieval} There are several corpora which feature direct question and answering sessions. These may involve general QA, such as in a press conference, or more task-specific lines of questioning, as to retrieve a specific set of information. 

The \textbf{Corpus of Professional Spoken American English} (CPSAE) \citep{barlow2000corpus} was constructed using a selection of transcripts of interactions occurring in professional settings. The corpus contains two million words involving over 400 speakers, recorded between 1994-1998. The CPASE has two main components. The first is a collection of transcripts (0.9 million words) of White House press conferences, which contains almost exclusively question and answer sessions, with some policy statements by politicians. The second component consists of transcripts (1.1 million words) of faculty meetings and committee meetings related to national tests that involve statements, discussions, and questions. The creation of the corpus was motivated by the desire to understand and model more formal uses of the English language.

As previously mentioned, the Dialog State Tracking Challenge (DSTC) consists of a series of datasets evaluated using a `state tracking' or `slot filling' metric. While the first 3 installments of this challenge had conversations between a human participant and a computer, \textbf{DSTC4} \citep{kim2015dialog} contains dialogues between humans. In particular, this dataset has 35 conversations with 21 hours of interactions between tourists and tour guides over Skype, discussing information on hotels, flights, and car rentals. Due to the small size of the dataset, researchers were encouraged to use transfer learning from other datasets in the DSTC in order to improve state tracking performance. This same training set is used for \textbf{DSTC5} \citep{kim2016fifth} as well. However, the goal of DSTC5 is to study multi-lingual speech-act prediction, and therefore it combines the DSTC4 dialogues plus a set of equivalent Chinese dialogs; evaluation is done on a holdout set of Chinese dialogues.

\paragraph{Miscellaneous} Lastly, there are several corpora which do not fall into any of the aforementioned categories, involving a range of tasks and situations.

The \textbf{IDIAP Wolf Corpus} \citep{hung2010idiap} is an audio-visual corpus containing natural conversational data of volunteers who took part in an adversarial role-playing game called `Werewolf'. Four groups of 8-12 people were recorded using headset microphones and synchronised video cameras, resulting in over 7 hours of conversational data. The novelty of this dataset is that the roles of other players are unknown to game participants, and some of the roles are deceptive in nature. Thus, there is a significant amount of lying that occurs during the game. Although specific instances of lying are not annotated, each speaker is labeled with their role in the game. In a dialogue setting, this could be useful for analyzing the differences in language when deception is being used.

The \textbf{SEMAINE Corpus} \citep{mckeown2010semaine} consists of 100 `emotionally coloured' conversations. Participants held conversations with an operator who adopted various roles designed to evoke
emotional reactions. These conversations were recorded with synchronous video and audio devices. Importantly, the operators' responses were stock phrases that were independent of the content of the user's utterances, and only dependent on the user's emotional state. This corpus motivates building dialogue systems with affective and emotional intelligence abilities, since the corpus does not exhibit the natural language understanding that normally occurs between human interlocutors.

The \textbf{Loqui Human-Human Dialogue Corpus} \citep{passonneau2014loqui} consists of annotated transcriptions of telephone interactions between patrons and librarians at New York City's Andrew Heiskell Braille \& Talking Book Library in 2006. It stands out as it has annotated discussion topics, question-answer pair links (adjacency pairs), dialogue acts, and frames (discourse units).

Similarly, the \textbf{The ICSI Meeting Recorder Dialog Act (MRDA) Corpus} \citep{shriberg2004icsi} has annotated dialogue acts, question-answer pair links (adjacency pairs), and dialogue hot spots\footnote{For more information on dialogue hot spots and how they relate to dialogue acts, see \citep{wrede2003relationship}.}. It consists of transcribed recordings of 75 ICSI meetings on several classes of topics including: the ICSI meeting recorder project itself, automatic speech recognition, natural language processing and neural theories of language, and discussions with the annotators for the project.

\subsubsection{Scripted Corpora}

A final category of spoken dialogue consists of conversations that have been pre-scripted for the purpose of being spoken later. We refer to datasets containing such conversations as `scripted corpora'. As discussed in Subsection \ref{subsec:corpora_from_fiction}, these datasets are distinct from spontaneous human-human conversations, as they inevitably contain fewer `filler' words and expressions that are common in spoken dialogue. However, they should not be confused with human-human written dialogues, as they are intended to sound like natural spoken conversations when read aloud by the participants. Furthermore, these scripted dialogues are required to be dramatic, as they are generally sourced from movies or TV shows.

There exist multiple scripted corpora based on movies and TV series. These can be sub-divided into two categories: corpora that provide the actual scripts (i.e.\@ the movie script or TV series script) where each utterance is tagged with the appropriate speaker, and those that only contain subtitles and consecutive utterances are not divided or labeled in any way. It is always preferable to have the speaker labels, but there is significantly more unlabeled subtitle data available, and both sources of information can be leveraged to build a dialogue system.

The \textbf{Movie DiC Corpus} \citep{banchs2012movie} is an example of the former case---it contains about 130,000 dialogues and 6 million words from movie scripts extracted from the Internet Movie Script Data Collection\footnote{\url{http://www.imsdb.com}}, carefully selected to cover a wide range of genres. These dialogues also come with context descriptions, as written in the script. One derivation based on this corpus is the \textbf{Movie Triples Dataset} \citep{2015arXiv150704808S}. 
There is also the \textbf{American Film Scripts Corpus} and \textbf{Film Scripts Online Corpus} which form the \textbf{Film Scripts Online Series Corpus}, which can be purchased \footnote{\url{http://alexanderstreet.com/products/film-scripts-online-series}}. The latter consists of a mix of British and American film scripts, while the former consists of solely American films.


The majority of these datasets consist mostly of raw scripts, which are not guaranteed to portray conversations between only two people. The dataset collected by \cite{nio2014conversation}, which we refer to as the \textbf{Filtered Movie Script Corpus}, takes over 1 million utterance-response pairs from web-based script resources and filters them down to 86,000 such pairs. The filtering method limits the extracted utterances to X-Y-X triples, where X is spoken by the same actor and each of the utterance share some semantic similarity. These triples are then decomposed into X-Y and Y-X pairs. Such filtering largely removes conversations with more than two speakers, which could be useful in some applications. Particularly, the filtering method helps to retain semantic context in the dialogue and keeps a back-and-forth conversational flow that is desired in training many dialogue systems.

The \textbf{Cornell Movie-Dialogue Corpus} \citep{Danescu-Niculescu-Mizil+Lee:11a} also has short conversations extracted from movie scripts. The distinguishing feature of this dataset is the amount of metadata available for each conversation: this includes movie metadata such as genre, release year, and IMDB rating, as well as character metadata such as gender and position on movie credits. Although this corpus contains 220,000 dialogue excerpts, it only contains 300,000 utterances; thus, many of the excerpts consist of single utterances.

The \textbf{Corpus of American Soap Operas} \citep{davies2012corpus} contains 100 million words in more than 22,000 transcripts of ten American TV-series soap operas from 2001 and 2012. Because it is based on soap operas it is qualitatively different from the Movie Dic Corpus, which contains movies in the action and horror genres. The corpus was collected to provide insights into colloquial American speech, as the vocabulary usage is quite different from the British National Corpus \citep{davies2012comparing}. Unfortunately, this corpus does not come with speaker labels.

Another corpus consisting of dialogues from TV shows is the \textbf{TVD Corpus} \citep{roy2014tvd}. This dataset consists of 191 movie transcripts from the comedy show \textit{The Big Bang Theory}, and the drama show \textit{Game of Thrones}, along with crowd-sourced text descriptions (brief episode summaries, longer episode outlines) and various types of metadata (speakers, shots, scenes). Text alignment algorithms are used to link descriptions and metadata to the appropriate sections of each script. For example, one might align an event description with all the utterances associated with that event in order to develop algorithms for locating specific events from raw dialogue, such as 'person X tries to convince person Y'.

Some work has been done in order to analyze character style from movie scripts. This is aided by a dataset collected by \cite{walker2012annotated} that we refer to as the \textbf{Character Style from Film Corpus}. This corpus was collected from the IMSDb archive, and is annotated for linguistic structures and character archetypes. Features, such as the sentiment behind the utterances, are automatically extracted and used to derive models of the characters in order to generate new utterances similar in style to those spoken by the character. Thus, this dataset could be useful for building dialogue personalization models.

There are two primary movie subtitle datasets: the \textbf{OpenSubtitles} \citep{tiedemann2012parallel} and the \textbf{SubTle Corpus} \citep{ameixa2013subtitles}. Both corpora are based on the OpenSubtitles website.\footnote{\url{http://www.opensubtitles.org}} The OpenSubtitles dataset is a giant collection of movie subtitles, containing over 1 billion words, whereas SubTle Corpus has been pre-processed in order to extract interaction-response pairs that can help dialogue systems deal with out-of-domain (OOD) interactions.

The \textbf{Corpus of English Dialogues 1560–-1760 (CED)} \citep{kyto2006guide} compiles dialogues from the mid-16th century until the mid-18th century. The sources vary from real trial transcripts to fiction dialogues. Due to the scripted nature of fictional dialogues and the fact that the majority of the corpus consists of fictional dialogue, we classify it here as such. The corpus is composed as follows: trial proceedings (285,660 words), witness depositions (172,940 words), drama comedy works (238,590 words), didactic works (236,640 words), prose fiction (223,890 words), and miscellaneous (25,970 words).

\subsection{Human-Human Written Corpora}\label{sec:hhwrit}

We proceed to survey corpora of conversations between humans in written form. As before, we sub-divide this section into \textit{spontaneous} and \textit{constrained} corpora, depending on whether there are restrictions on the topic of conversation. However, we make a further distinction between \textit{forum}, \textit{micro-blogging}, and \textit{chat} corpora.

Forum corpora consist of conversations on forum-based websites such as Reddit\footnote{\url{http://www.reddit.com}} where users can make posts, and other users can make comments or replies to said post. In some cases, comments can be nested indefinitely, as users make replies to previous replies. Utterances in forum corpora tend to be longer, and there is no restriction on the number of participants in a discussion. On the other hand, conversations on micro-blogging websites such as Twitter\footnote{\url{http://www.twitter.com}} tend to have very short utterances as there is an upper bound on the number of characters permitted in each message. As a result, these tend to exhibit highly colloquial language with many abbreviations. The identifying feature of chat corpora is that the conversations take place in real-time between users. Thus, these conversations share more similarities with spoken dialogue between humans, such as common grounding phenomena.

\subsubsection{Spontaneous Written Corpora}

We begin with written corpora where the topic of conversation is not pre-specified. Such is the case for the \textbf{NPS Internet Chatroom Conversations Corpus} \citep{forsyth2007lexical}, which consists of 10,567 English utterances gathered from age-specific chat rooms of various online chat services from October and November of 2006. Each utterance was annotated with part-of-speech and dialogue act information; the correctness of this was verified manually. The NPS Internet Chatroom Conversations Corpus was one of the first corpora of computer-mediated communication (CMC), and it was intended for various NLP applications such as conversation thread topic detection, author profiling, entity identification, and social network analysis.

Several corpora of spontaneous micro-blogging conversations have been collected, such as the \textbf{Twitter Corpus} from \shortcite{Ritter:2010:UMT:1857999.1858019}, which contains 1.3 million post-reply pairs extracted from Twitter. The corpus was originally constructed to aid in the production of unsupervised approaches to modeling dialogue acts. Larger Twitter corpora have been collected. The \textbf{Twitter Triples Corpus} \citep{sordoni2015aneural} is one such example, with a described original dataset of 127 million context-message-response triples, but only a small labeled subset of this corpus has been released. Specifically, the released labeled subset contains 4,232 pairs that scored an average of greater than 4 on the Likert scale by crowdsourced evaluators for quality of the response to the context-message pair. Similarly, large micro-blogging corpora such as the \textbf{Sina Weibo Corpus} \citep{shang2015neural}, which contains 4.5 million post-reply pairs, have been collected; however, the authors have not yet been made publicly available. We do not include the Sina Weibo Corpus (and its derivatives) in the tables in this section, as they are not primarily in English.


The \textbf{Usenet Corpus} \citep{shaoul2009usenet} is a gigantic collection of public Usenet postings\footnote{\url{http://www.usenet.net}} containing over 7 billion words from October 2005 to January 2011. Usenet was a distributed discussion system established in 1980 where participants could post articles to one of 47,860 `newsgroup' categories. It is seen as the precursor to many current Internet forums. The corpus derived from these posts has been used for research in collaborative filtering \citep{konstan1997grouplens} and role detection \citep{fisher2006you}.

The \textbf{NUS SMS Corpus} \citep{chen2013creating} consists of conversations carried out over mobile phone SMS messages between two users. While the original purpose of the dataset was to improve predictive text entry when mobile phones still mapped multiple letters to a single number, aided by video and timing analysis of users entering their messages it could equally be used for analysis of informal dialogue. Unfortunately, the corpus does not consist of dialogues, but rather single SMS messages. SMS messages are similar in style to Twitter, in that they use many abbreviations and acronyms.

Currently, one of the most popular forum-based websites is Reddit\footnote{\url{http://www.reddit.com}} where users can create discussions and post comments in various sub-forums called `subreddits'. Each subreddit addresses its own particular topic. Over 1.7 billion of these comments have been collected in the \textbf{Reddit Corpus}.\footnote{\url{https://www.reddit.com/r/datasets/comments/3bxlg7/i\_have\_every\_publicly\_available\_reddit\_comment/}} Each comment is labeled with the author, score (rating from other users), and position in the comment tree; the position is important as it determines which comment is being replied to. Although researchers have not yet investigated dialogue problems using this Reddit discussion corpus, the sheer size of the dataset renders it an interesting candidate for transfer learning. Additionally, researchers have used smaller collections of Reddit discussions for broad discourse classification. \citep{ray4183analysis}.

Some more curated versions of the Reddit dataset have been collected. The \textbf{Reddit Domestic Abuse Corpus} \citep{ray4183analysis} consists of Reddit posts and comments taken from either subreddits specific to domestic abuse, or from subreddits representing casual conversations, advice, and general anxiety or anger. The motivation is to build classifiers that can detect occurrences of domestic abuse in other areas, which could provide insights into the prevalence and consequences of these situations. These conversations have been pre-processed with lower-casing, lemmatizing, and removal of stopwords, and semantic role labels are provided.


\subsubsection{Constrained Written Corpora}

There are also several written corpora where users are limited in terms of topics of conversation. For example, the \textbf{Settlers of Catan Corpus} \citep{afantenos2012developing} contains logs of 40 games of `Settlers of Catan', with about 80,000 total labeled utterances. The game is played with up to 4 players, and is predicated on trading certain goods between players. The goal of the game is to be the first player to achieve a pre-specified number of points. Therefore, the game is adversarial in nature, and can be used to analyze situations of strategic conversation where the agents have diverging motives.

Another corpus that deals with game playing is the \textbf{Cards Corpus} \citep{djalali2012corpus}, which consists of 1,266 transcripts of conversations between players playing a game in the `Cards world'. This world is a simple 2-D environment where players collaborate to collect cards. The goal of the game is to collect six cards of a particular suit (cards in the environment are only visible to a player when they are near the location of that player), or to determine that this goal is impossible in the environment. The catch is that each player can only hold 3 cards, thus players must collaborate in order to achieve the goal. Further, each player's location is hidden to the other player, and there are a fixed number of non-chatting moves. Thus, players must use the chat to formulate a plan, rather than exhaustively exploring the environment themselves. The dataset has been further annotated by \cite{potts2012goal} to collect all locative question-answer pairs (i.e.\@ all questions of the form `Where are you?').

The \textbf{Agreement by Create Debaters Corpus} \citep{rosenthal2015couldn}, the \textbf{Agreement in Wikipedia Talk Pages Corpus} \citep{andreas2012annotating} and the \textbf{Internet Argument Corpus} \citep{abbott2016internet} all cover dialogs with annotations measuring levels of agreement or disagreement in responses to posts in various media. The \textbf{Agreement by Create Debaters Corpus} and the \textbf{Agreement in Wikipedia Talk Pages Corpus} both are formatted in the same way. Post-reply pairs are annotated with whether they are in agreement or disagreement, as well as the type of agreement they are in if applicable (e.g. paraphrasing). The difference between the two corpora is the source: the former is collected from Create Debate forums and the latter from a mix of Wikipedia Discussion pages and LiveJournal postings. The \textbf{Internet Argument Corpus} (IAC) \citep{walker2012corpus} is a forum-based corpus with 390,000 posts on 11,000 discussion topics. Each topic is controversial in nature, including subjects such as evolution, gay marriage and climate change; users participate by sharing their opinions on one of these topics. Posts-reply pairs have been labeled as being either in agreement or disagreement, and sarcasm ratings are given to each post.

Another source of constrained text-based corpora are chat-room environments. Such a set-up forms the basis of the \textbf{MPC Corpus} \citep{shaikh2010mpc}, which consists of 14 multi-party dialogue sessions of approximately 90 minutes each. In some cases, discussion topics were constrained to be about certain political stances, or mock committees for choosing job candidates.
An interesting feature is that different participants are given different roles---leader, disruptor, and consensus builder---with only a general outline of their goals in the conversation. Thus, this dataset could be used to model social phenomena such as agenda
control, influence, and leadership in on-line interactions.

The largest written corpus with a constrained topic is the recently released \textbf{Ubuntu Dialogue Corpus} \citep{lowe2015ubuntu}, which has almost 1 million dialogues of 3 turns or more, and 100 million words. It is related to the former \textbf{Ubuntu Chat Corpus} \citep{uthus2013ubuntu}. Both corpora were scraped from the Ubuntu IRC channel logs.\footnote{\url{http://irclogs.ubuntu.com}} On this channel, users can log in and ask a question about a problem they are having with Ubuntu; these questions are answered by other users. Although the chat room allows everyone to chat with each other in a multi-party setting, the Ubuntu Dialogue Corpus uses a series of heuristics to disentangle it into dyadic dialogue. The technical nature and size of this corpus lends itself particularly well to applications in technical support.

Other corpora have been extracted from IRC chat logs. The \textbf{IRC Corpus} \citep{elsner2008you} contains approximately 50 hours of chat, with an estimated 20,000 utterances from the Linux channel on IRC, complete with the posting times. Therefore, this dataset consists of similarly technical conversations to the Ubuntu Corpus, with the occasional social chat. The purpose of this dataset was to investigate approaches for conversation disentanglement; given a multi-party chat room, one attempts to recover the individual conversations of which it is composed. For this purpose, there are approximately 1,500 utterances with annotated ground-truth conversations.

More recent efforts have combined traditional conversational corpora with question answering and recommendation datasets in order to facilitate the construction of goal-driven dialogue systems. Such is the case for the \textbf{Movie Dialog Dataset}~\citep{dodge2015evaluating}. There are four tasks that the authors propose as a prerequisite for a working dialogue system: question answering, recommendation, question answering with recommendation, and casual conversation. The Movie Dialog dataset consists of four sub-datasets used for training models to complete these tasks: a QA dataset from the Open Movie Database (OMDb)\footnote{\url{http://en.omdb.org}} of 116k examples with accompanying movie and actor metadata in the form of knowledge triples; a recommendation dataset from MovieLens\footnote{\url{http://movielens.org}} with 110k users and 1M questions; a combined recommendation and QA dataset with 1M conversations of 6 turns each; and a discussion dataset from Reddit's movie subreddit. The former is evaluated using recall metrics in a manner similar to \citet{lowe2015ubuntu}. It should be noted that, other than the Reddit dataset, the dialogues in the sub-datasets are simulated QA pairs, where each response corresponds to a list of entities from the knowledge base.

\section{Discussion}
We conclude by discussing a number of general issues related to the development and evaluation of data-driven dialogue systems. We also discuss alternative sources of information, user personalization, and automatic evaluation methods.

\subsection{Challenges of Learning from Large Datasets}
\label{sec:bigdata}

Recently, several large-scale dialogue datasets have been proposed in order to train data-driven dialogue systems; the Twitter Corpus \citep{Ritter:2010:UMT:1857999.1858019} and the Ubuntu Dialogue corpus \citep{lowe2015ubuntu} are two examples.
In this section, we discuss the benefits and drawbacks of these datasets based on our experience using them for building data-driven models.
Unlike the previous section, we now focus explicitly on aspects of high relevance for using these datasets for learning dialogue strategies.

\subsubsection{The Twitter Corpus} 

The Twitter Corpus consists of a series of conversations extracted from tweets. While the dataset is large and general-purpose, the micro-blogging nature of the source material leads to several drawbacks for building conversational dialogue agents.
However, some of these drawbacks do not apply if the end goal is to build an agent that interacts with users on the Twitter platform.

The Twitter Corpus has an enormous amount of typos, slang, and abbreviations. Due to the 140-character limit, tweets are often very short and compressed.
In addition, users frequently use Twitter-specific devices such as hashtags. 
Unless one is building a dialogue agent specifically for Twitter, it is often not desirable to have a chatbot use hashtags and excessive abbreviations as it is not reflective of how humans converse in other environments.
This also results in a significant increase in the word vocabulary required for dialogue systems trained at the word level.
As such, it is not surprising that character-level models have shown promising results on Twitter~\citep{dhingra2016tweet2vec}.

Twitter conversations often contain various kinds of verbal role-playing and imaginative actions similar to stage directions in theater plays (e.g.\@ instead of writing \textit{``goodbye''}, a user might write ``*waves goodbye and leaves*'').
These conversations are very different from the majority of text-based chats.
Therefore, dialogue models trained on this dataset are often able to provide interesting and accurate responses to contexts involving role-playing and imaginative actions \citep{serban2016hierarchical}. 

Another challenge posed by Twitter is that Twitter conversations often refer to recent public events outside the conversation.
In order to learn effective responses for such conversations, a dialogue agent must infer the news event under discussion by referencing some form of external knowledge base.
This would appear to be a particularly difficult task.


\subsubsection{The Ubuntu Dialogue Corpus}

The Ubuntu Dialogue Corpus is one of the largest, publicly available datasets containing technical support dialogues.
Due to the commercial importance of such systems, the dataset has attracted significant attention.\footnote{Most of the largest technical support datasets are based on commercial technical support channels, which are proprietary and never released to the public for privacy reasons.}
Thus, the Ubuntu Dialogue Corpus presents the opportunity for anyone to train large-scale data-driven technical support dialogue systems.

Despite this, there are several problems when training data-driven dialogue models on the Ubuntu Dialogue Corpus due to the nature of the data.
First, since the corpus comes from a multi-party IRC channel, it needs to be disentangled into separate dialogues.
This disentanglement process is noisy, and errors inevitably arise.
The most frequent error is when a missing utterance in the dialogue is not picked up by the extraction procedure (e.g.\@ an utterance from the original multi-party chat was not added to the disentangled dialogue).
As a result, for a substantial amount of conversations, it is difficult to follow the topic.
In particular, this means that some of the Next Utterance Classification (NUC) examples, where models must select the correct next response from a list of candidates, are either difficult or impossible for models to predict.

Another problem arises from the lack of annotations and labels.
Since users try to solve their technical problems, it is perhaps best to build models under a goal-driven dialogue framework, where a dialogue system has to maximize the probability that it will solve the user's problem at the end of the conversation.
However, there are no reward labels available.
Thus, it is difficult to model the dataset in a goal-driven dialogue framework.
Future work may alleviate this by constructing automatic methods of determining whether a user in a particular conversation solved their problem.

A particular challenge of the Ubuntu Dialogue Corpus is the large number of out-of-vocabulary words, including
many technical words related to the Ubuntu operating system, such as commands, software packages, websites, etc.
Since these words occur rarely in the dataset, it is difficult to learn their meaning directly from the dataset --- for example, it is difficult to obtain meaningful distributed, real-valued vector representations for neural network-based dialogue models.
This is further exacerbated by the large number of users who use different nomenclature, acronyms, and speaking styles, and the many typos in the dataset. Thus, the linguistic diversity of the corpus is large.

A final challenge of the dataset is the necessity for additional knowledge related to Ubuntu in order to accurately generate or predict the next response in a conversation.
We hypothesize that this knowledge is crucial for a system trained on the Ubuntu Dialogue Corpus to be effective in practice, as often solutions to technical problems change over time as new versions of the operating system become available.
Thus, an effective dialogue system must learn to combine up-to-date technical information with an understanding of natural language dialogue in order to solve the users' problems.
We will discuss the use of external knowledge in more detail in Section \ref{sec:knowledge}.

While these challenges make it difficult to build data-driven dialogue systems, it also presents an important research opportunity.
Current data-driven dialogue systems perform rather poorly in terms of generating utterances that are coherent and on-topic \citep{serban2016multiresolution}.
As such, there is significant room for improvement on these models.


\subsection{Transfer Learning Between Datasets}
While it is not always feasible to obtain large corpora for every new application, the use of other related datasets can effectively bootstrap the learning process.
In several branches of machine learning, and in particular in deep learning, the use of related datasets in pre-training the model is an effective method of scaling up to complex environments~\citep{erhan2010jmlr,kumar2015ask}.

To build open-domain dialogue systems, it is arguably necessary to move beyond domain-specific datasets.
Instead, like humans, dialogue systems may have to be trained on multiple data sources for solving multiple tasks.
To leverage statistical efficiency, it may be necessary to first use unsupervised learning---as opposed to supervised learning or offline reinforcement learning, which typically only provide a sparse scalar feedback signal for each phrase or sequence of phrases---and then fine-tune models based on human feedback.
Researchers have already proposed various ways of applying transfer learning to build data-driven dialogue systems, ranging from learning separate sub-components of the dialogue system (e.g.\@ intent and dialogue act classification) to learning the entire dialogue system (e.g.\@ in an unsupervised or reinforcement learning framework) using transfer learning \citep{fabbrizio_bootstrappingspoken,forguesbootstrapping,serban2015textbased,2015arXiv150704808S,lowe2015ubuntu,vandyke2015multi,wen2016multi,gavsic2016dialogue,mo2016personalizing,genevay2016transfer,chen2016zero}

\subsection{Topic-oriented \& Goal-driven Datasets}
Tables~\ref{tab:hc}--\ref{tab:hh_written} list the topics of available datasets. 
Several of the human-human datasets are denoted as having \textit{casual} or \textit{unrestricted} topics.
In contrast, most human-machine datasets focus on specific, narrow topics.
It is useful to keep this distinction between \textit{restricted} and \textit{unrestricted} topics in mind, as goal-driven dialogue systems --- which typically have a well-defined measure of performance related to task completion --- are usually developed in the former setting.
In some cases, the line between these two types of datasets blurs.
For example, in the case of conversations occurring between players of an online game~\citep{afantenos2012developing}, the outcome of the game is determined by how participants play in the game environment, not by their conversation.
In this case, some conversations may have a direct impact on a player's performance in the game, some conversations may be related to the game but irrelevant to the goal (e.g.\@ commentary on past events) and some conversations may be completly unrelated to the game.

\subsection{Incorporating longer memories}
Recently, significant progress has been made towards incorporating a form of external memory into various neural-network architectures for sequence modeling.
Models such as Memory Networks~\citep{weston15,sukhbaatar15} and Neural Turing Machines (NTM)~\citep{graves14} store some part of their input in a memory, which is then reasoned over in order to perform a variety of sequence to sequence tasks. These vary from simple problems, such as sequence copying, to more complex problems, such as question answering and machine translation. Although none of these models are explicitly designed to address dialogue problems, the extension by \shortcite{kumar2015ask} to Dynamic Memory Networks specifically differentiates between episodic and semantic memory. In this case, the episodic memory is the same as the memory used in the traditional Memory Networks paper that is extracted from the input, while  the semantic memory refers to knowledge sources that are fixed for all inputs. The model is shown to work for a variety of NLP tasks, and it is not difficult to envision an application to dialogue utterance generation where the semantic memory is the desired external knowledge source.

\subsection{Incorporating External Knowledge}
\label{sec:knowledge}
Another interesting research direction is the incorporation of external knowledge sources in order to inform the response to be generated. Using external information is of great importance to dialogues systems, particularly in the goal-driven setting.
Even non-goal-driven dialogue systems designed to simply entertain the user could benefit from leveraging external information, such as current news articles or movie reviews, in order to better converse about real-world events. This may be particularly useful in data-sparse domains, where there is not enough dialogue training data to reliably learn a response that is appropriate for each input utterance, or in domains that evolve quickly over time.

\subsubsection{Structured External Knowledge}
In traditional goal-driven dialogue systems~\citep{levin1997stochastic}, where the goal is to provide information to the user, there is already extensive use of external knowledge sources. For example, in the Let's Go! dialogue system~\citep{raux2005let}, the user requests information about various bus arrival and departure times. Thus, a critical input to the model is the actual bus schedule, which is used in order to generate the system's utterances. Another example is the dialogue system described by \shortcite{noth2004experiences}, which helps users find movie information by utilizing movie showtimes from different cinemas.
Such examples are abundant both in the literature and in practice. 
Although these models make use of external knowledge, the knowledge sources in these cases are highly structured and are only used to place hard constraints on the possible states of an utterance to be generated. They are essentially contained in relational databases or structured ontologies, and are only used to provide a deterministic mapping from the dialogue states extracted from an input user utterance to the dialogue system state or the generated response.

Complementary to domain-specific databases and ontologies are the general natural language processing databases and tools. These include lexical databases such as WordNet \citep{miller1995wordnet}, which contains lexical relationships between words for over a hundred thousand words, VerbNet \citep{schuler2005verbnet} which contains lexical relations between verbs, and FrameNet \citep{RuppenhoferEtAl2006:ExtTeoR.dPractice}, which contains 'word senses' for over ten thousand words along with examples of each word sense. In addition, there exist several natural language processing tools such as part of speech taggers, word category classifiers, word embedding models, named entity recognition models, co-reference resolution models, semantic role labeling models, semantic similarity models and sentiment analysis models \citep{manning1999foundations,jurafsky2008speech,mikolov2013distributed,gurevych2004semantic,lin2011all} that may be used by the Natural Language Interpreter to extract meaning from human utterances. Since these tools are typically built upon texts and annotations created by humans, using them inside a dialogue system can be interpreted as a form of structured transfer learning, where the relationships or labels learned from the original natural language processing corpus provide additional information to the dialogue system and improve generalization of the system.

\subsubsection{Unstructured External Knowledge}
Complementary sources of information can be found in unstructured knowledge sources, such as online encyclopedias (Wikipedia \citep{denoyer2007wikipedia})
as well as domain-specific sources~\citep{lowe2015incorporating}.
It is beyond the scope of this paper to review all possible ways that these unstructured knowledge sources have or could be used in conjunction with a data-driven dialogue system.
However, we note that this is likely to be a fruitful research area. 

\subsection{Personalized dialogue agents}
When conversing, humans often adapt to their interlocutor to facilitate understanding, and thus improve conversational efficiency and satisfaction.
Attaining human-level performance with dialogue agents may well require personalization, i.e.\@ models that are aware and capable of adapting to their intelocutor.
Such capabilities could increase the effectiveness and naturalness of generated dialogues \citep{lucas2009,su2013dialogue}.
We see personalization of dialogue systems as an important task, which so far has not received much attention.
There has been initial efforts on user-specific models which could be adapted to work in combination with the dialogue models presented in this survey \citep{lucas2009,lin2011,Pargellis2004}.
There has also been interesting work on character modeling in movies \citep{walker2011,li2016persona,mo2016personalizing}.
There is significant potential to learn user models as part of dialogue models. The large datasets presented in this paper, some of which provide multiple dialogues per user, may enable the development of such models.

\subsection{Evaluation metrics}

One of the most challenging aspects of constructing dialogue systems lies in their evaluation.
While the end goal is to deploy the dialogue system in an application setting and receive real human feedback, getting to this stage is time consuming and expensive.
Often it is also necessary to optimize performance on a pseudo-performance metric prior to release.
This is particularly true if a dialogue model has many hyper-parameters to be optimized---it is infeasible to run user experiments for every parameter setting in a grid search.
Although crowdsourcing platforms, such as Amazon Mechanical Turk, can be used for some user testing \citep{jurcicek2011real}, evaluations using paid subjects can also lead to biased results~\citep{young2013pomdp}.
Ideally, we would have some automated metrics for calculating a score for each model, and only involve human evaluators once the best model has been chosen with reasonable confidence.

The evaluation problem also arises for non-goal-driven dialogue systems.
Here, researchers have focused mainly on the output of the response generation module.
Evaluation of such non-goal-driven dialogue systems can be traced back to the Turing test \citep{turing1950computing}, where human judges communicate with both computer programs and other humans over a chat terminal without knowing each other's true identity.
The goal of the judges was to identify the humans and computer programs under the assumption that a program indistinguishable from a real human being must be intelligent.
However, this setup has been criticized extensively with numerous researchers proposing alternative evaluation procedures \citep{cohen2005if}.
More recently, researchers have turned to analyzing the collected dialogues produced after they are finished \citep{GalleyBSJAQMGD15,pietquin2013survey,shawar2007different,schatzmann2005quantitative}.

Even when human evaluators are available, it is often difficult to choose a set of informative and consistent criteria that can be used to judge an utterance generated by a dialogue system.
For example, one might ask the evaluator to rate the utterance on vague notions such as `appropriateness' and `naturalness', or to try to differentiate between utterances generated by the system and those generated by actual humans \citep{vinyals2015neural}.
\shortcite{schatzmann2005quantitative} suggest two aspects that need to be evaluated for all response generation systems (as well as user simulation models): 1) if the model can generate human-like output, and 2) if the model can reproduce the variety of user behaviour found in corpus.
But we lack a definitive framework for such evaluations.

We complete this discussion by summarizing different approaches to the automatic evaluation problem as they relate to these objectives.

\subsubsection{Automatic Evaluation Metrics for goal-driven Dialogue Systems}

User evaluation of goal-driven dialogue systems typically focuses on goal-related performance criteria, such as goal completion rate, dialogue length, and user satisfaction \citep{walker1997paradise,schatzmann2005quantitative}.
These were originally evaluated by human users interacting with the dialogue system, but more recently researchers have also begun to use third-party annotators for evaluating recorded dialogues \citep{yang2010collection}.
Due to their simplicity, the vast majority of hand-crafted task-oriented dialogue systems have been solely evaluated in this way. However, when using machine learning algorithms to train on large-scale corpora, automatic optimization criteria are required. The challenge with evaluating goal-driven dialogue systems without human intervention is that the process necessarily requires multiple steps---it is difficult to determine if a task has been solved from a single utterance-response pair from a conversation. Thus, simulated data is often generated by a \textit{user simulator} \citep{eckert1997user,schatzmann2007agenda,jung2009data, georgila2006user,pietquin2013survey}.
Given a sufficiently accurate user simulation model, an interaction between the dialogue system and the user can be simulated from which it is possible to deduce the desired metrics, such as goal completion rate.
Significant effort has been made to render the simulated data as realistic as possible, by modeling user intentions. Evaluation of such simulation methods has already been conducted~\citep{schatzmann2005quantitative}. However, generating realistic user simulation models remains an open problem.

\subsubsection{Automatic Evaluation Metrics for Non-goal-driven Dialogue Systems}

Evaluation of non-goal-driven dialogue systems, whether by automatic means or user studies, remains a difficult challenge.

\textbf{Word Overlap Metrics}. One approach is to borrow evaluation metrics from other NLP tasks such as machine translation, which uses BLEU \citep{papineni2002bleu} and METEOR \citep{banerjee2005meteor} scores.
These metrics have been used to compare responses generated by a learned dialogue strategy to the actual next utterance in the conversation, conditioned on a dialogue context \citep{sordoni2015aneural}. 
While BLEU scores have been shown to correlate with human judgements for machine translation \citep{papineni2002bleu}, their effectiveness for automatically assessing dialogue response generation is unclear.
There are several issues to consider: given the context of a conversation, there often exists a large number of possible responses that `fit' into the dialogue.
Thus, the response generated by a dialogue system could be entirely reasonable, yet it may have no words in common with the actual next utterance.
In this case, the BLEU score would be very low, but would not accurately reflect the strength of the model.
Indeed, even humans who are tasked with predicting the next utterance of a conversation achieve relatively low BLEU scores~\citep{sordoni2015aneural}.
Although the METEOR metric takes into account synonyms and morphological variants of words in the candidate response, it still suffers from the aforementioned problems. 
In a sense, these measurements only satisfy one direction of Schatzmann's criteria: high BLEU and METEOR scores imply that the model is generating human-like output, but the model may still not reproduce the variety of user behaviour found in corpus.
Furthermore, such metrics will only accurately reflect the performance of the dialogue system if given a large number of candidate responses for each given context.

\textbf{Next Utterance Classification}. Alternatively, one can narrow the number of possible responses to a small, pre-defined list, and ask the model to select the most appropriate response from this list.
The list includes the actual next response of the conversation (the desired prediction), and the other entries (false positives) are sampled from elsewhere in the corpus~\citep{lowe2016evaluation,lowe2015ubuntu}.
This \textit{next utterance classification} (NUC) task is derived from the recall and precision metrics for information-retrieval-based approaches.
There are several attractive properties of this metric: it is easy to interpret, and the difficulty can be adjusted by changing the number of false responses.
However, there are drawbacks. In particular, since the other candidate answers are sampled from elsewhere in the corpus, there is a chance that these also represent reasonable responses given the context.
This can be alleviated to some extent by reporting Recall@k measures, i.e.\@ whether the correct response is found in the k responses with the highest rankings according to the model.
Although current models evaluated using NUC are trained explicitly to maximize the performance  on this metric by minimizing the cross-entropy between context-response pairs \citep{lowe2015ubuntu, kadlec2015improved}, the metric could also be used to evaluate a probabilistic generative model trained to output full utterances.

\textbf{Word Perplexity}. Another metric proposed to evaluate probabilistic language models  \citep{bengio2003neural, mikolov2010recurrent} that has seen significant recent use for evaluating end-to-end dialogue systems is \textit{word perplexity} \citep{pietquin2013survey, 2015arXiv150704808S}.
Perplexity explicitly measures the probability that the model will generate the ground truth next utterance given some context of the conversation. This is particularly appealing for dialogue, as the distribution over words in the next utterance can be highly multi-modal (i.e.\@ many possible responses). 
A re-weighted perplexity metric has also been proposed where stop-words, punctuation, and end-of-utterance tokens are removed before evaluating to focus on the semantic content of the phrase \citep{2015arXiv150704808S}. Both word perplexity, as well as utterance-level recall and precision outlined above, satisfy Schatzmann's evaluation criteria, since scoring high on these would require the model to produce human-like output and to reproduce most types of conversations in the corpus.

\textbf{Response Diversity}. Recent non-goal-driven dialogue systems based on neural networks have had problems generating diverse responses \citep{2015arXiv150704808S}.
\citep{li2015diversity} recently introduced two new metrics, \textit{distinct-1} and \textit{distinct-2}, which respectively measure the number of distinct unigrams and bigrams of the generated responses. Although these fail to satisfy either of Schatzmann's criteria, they may still be useful in combination with other metrics, such as BLEU, NUC or word perplexity.



\section{Conclusion}
There is strong evidence that over the next few years, dialogue research will quickly move towards large-scale data-driven model approaches. In particular, as is the case for other language-related applications such as speech recognition, machine translation and information retrieval, these approaches will likely come in the form of end-to-end trainable systems. This paper provides an extensive survey of currently available datasets suitable for research, development, and evaluation of such data-driven dialogue systems. 

In addition to presenting the datasets, we provide a detailed discussion of several of the issues related to the use of datasets in dialogue system research. Several potential directions are highlighted, such as transfer learning and incorporation of external knowledge, which may lead to scalable solutions for end-to-end training of conversational agents.

\subsection*{Acknowledgements}

The authors gratefully acknowledge financial support by the Samsung Advanced Institute of Technology (SAIT), the Natural Sciences and Engineering Research Council of Canada (NSERC), the Canada Research Chairs, the Canadian Institute for Advanced Research (CIFAR) and Compute Canada. Early versions of the manuscript benefited greatly from the proofreading of Melanie Lyman-Abramovitch, and later versions were extensively revised by Genevieve Fried and Nicolas Angelard-Gontier. The authors also thank Nissan Pow, Michael Noseworthy, Chia-Wei Liu, Gabriel Forgues, Alessandro Sordoni, Yoshua Bengio and Aaron Courville for helpful discussions.

{\footnotesize \bibliography{ref}}

\newpage

\appendix
\section{Learning from Dialogue Corpora}
In this appendix section, we review some existing computational architectures suitable for learning dialogue strategies directly from data.  The goal is not to provide full technical details on the methods available to achieve this --- though we provide appropriate citations for the interested reader --- but rather to illustrate concretely how the datasets described above can, and have, been used in different dialogue learning efforts.
As such, we limit this review to a small set of existing work.

\subsection{Data Pre-processing}

Before applying machine learning methods to a dialogue corpus, it is common practice to perform some form of pre-processing.
The aim of pre-processing is to standardize a dataset with minimal loss of information. This can reduce data scarcity, and eventually make it easier for models to learn from the dataset. In natural language processing, it is commonly acknowledged that pre-processing can have a significant effect on the results of the natural language processing system---the same observation holds for dialogue. 
Although the specific procedure for pre-processing is task- and data-dependent, in this section we highlight a few common approaches, in order to give a general idea of where pre-processing can be effective for dialogue systems.

Pre-processing is often used to remove anomalies in the data. For text-based corpora, this can include removing acronyms, slang, misspellings and phonemicization (e.g.\@ where words are written according to their pronunciation instead of their correct spelling). For some models, such as the generative dialogue models discussed later, tokenization (e.g.\@ defining the smallest unit of input) is also critical. 
In datasets collected from mobile text, forum, microblog and chat-based settings, it is common to observe a significant number of acronyms, abbreviations, and phonemicizations that are specific to the topic and userbase \citep{clark2003pre}. 
Although there is no widely accepted standard for handling such occurrences, many NLP systems incorporate some form of pre-processing to normalize these entries \citep{kaufmann2010syntactic,aw2006phrase,clark2003pre}. For example, there are look-up tables, such as the IRC Beginner List\footnote{\url{http://www.ircbeginner.com/ircinfo/abbreviations.html}}, which can be used to translate the most common acronyms and slang into standard English. Another common strategy is to use stemming and lemmatization to replace many words with a single item (e.g. \emph{walking} and \emph{walker} both replaced by \emph{walk}).  Of course, depending on the task at hand and the corpus size, an option is also to leave the acronyms and phonemicized words as they are.

In our experience, almost all dialogue datasets contain some amount of spelling errors. By correcting these, we expect to reduce data sparsity. This can be done by using automatic spelling correctors. However, it is important to inspect their effectiveness. For example, for movie scripts, \shortcite{2015arXiv150704808S} found that automatic spelling correctors introduced more spelling errors than they corrected, and a better strategy was to use Wikipedia's most commonly misspelled words\footnote{\url{https://en.wikipedia.org/wiki/Commonly\_misspelled\_English\_words}} to lookup and replace potential spelling errors. Transcribed spoken language corpora often include many non-words in their transcriptions (e.g.\@ uh, oh). Depending on whether or not these provide additional information to the dialogue system, researchers may also want to remove these words by using automatic spelling correctors.


\subsection{Segmenting Speakers and Conversations}

Some dialogue corpora, such as those based on movie subtitles, come without explicit speaker segmentation. 
However, it is often possible to estimate the speaker segmentation, which is useful to build a model of a given speaker---as compared to a model of the conversation as a whole. For text-based corpora, \shortcite{serban2015textbased} have recently proposed the use of recurrent neural networks to estimate turn-taking and speaker labels in movie scripts with promising results.

In the speech recognition literature, this is the subtask of speaker diarisation \citep{miro2012speaker,tranter2006overview}. 
When the audio stream of the speech is available, the segmentation is quite accurate with classification error rates as low as $5\%$. 

A strategy sometimes used for segmentation of spoken dialogues is based on labelling a small subset of the corpus, known as the gold corpus, and training a specific segmentation model based on this. The remaining corpus is then segmented iteratively according to the segmentation model, after which the gold corpus is expanded with the most confident segmentations and the segmentation model is retrained. This process is sometimes known as embedded training, and is widely used in other speech recognition tasks \citep{jurafsky2008speech}. It appears to work well in practice, but has the disadvantage that the interpretation of the label can drift. Naturally, this approach can be applied to text dialogues as well in a straightforward manner.


In certain corpora, such as those based on chat channels or extracted from movie subtitles, many conversations occur in sequence. In some cases, there are no labels partitioning the beginning and end of separate conversations. Similarly, certain corpora with multiple speakers, such as corpora based on chat channels, contain several conversations occurring in parallel (e.g.\@ simultaneously) but do not contain any segmentation separating these conversations. This makes it hard to learn a meaningful model from such conversations, because they do not represent consistent speakers or coherent semantic topics.

To leverage such data towards learning individual conversations, researchers have proposed methods to automatically estimate segmentations of conversations \citep{lowe2015ubuntu,nio2014developing}. Former solutions were mostly based on hand-crafted rules and seemed to work well upon manual inspection. For chat forums, one solution involves thresholding the beginning and end of conversations based on time (e.g. delay of more than $x$ minutes between utterances), and eliminating speakers from the conversation unless they are referred to explicitly by other speakers \citep{lowe2015ubuntu}. More advanced techniques involve maximum-entropy classifiers, which leverage the content of the utterances in addition to the discourse structure and timing information \citep{elsner2008you}. For movie scripts, researchers have proposed the use of simple information-retrieval similarity measures, such as cosine similarity, to identify conversations \citep{nio2014developing}. Based on the their performance on estimating turn-taking and speaker labels, recurrent neural networks also hold promise for segmenting conversations \citep{serban2015textbased}.



\subsection{Discriminative Model Architectures}
As discussed in Subsection \ref{subsec:discriminative_models}, discriminative models aim to predict certain labels or annotations manually associated with a portion of a dialogue. For example, a discriminative model might be trained to predict the intent of a person in a dialogue, or the topic, or a specific piece of information.

In the following subsections, we discuss research directions where discriminative models have been developed to solve dialogue-related tasks.\footnote{It is important to note that although discriminative models have been favored to model supervised problems in the dialogue-system literature, in principle generative models ($P(X,Y)$) instead of discriminative models ($P(Y|X)$) could be used.} This is primarily meant to review and contrast the work from a data-driven learning perspective.

\subsubsection{Dialogue Act Classification and Dialogue Topic Spotting}

Here we consider the simple task known as dialogue act classification (or dialogue move recognition). In this task, the goal is to classify a user utterance, independent of the rest of the conversation, as one out of $K$ dialogue acts: $P(A \mid U)$, where $A$ is the discrete variable representing the dialogue act and $U$ is the user's utterance. This falls under the general umbrella of text classification tasks, though its application is specific to dialogue. Like the dialogue state tracker model, a dialogue act classification model could be plugged into a dialogue system as an additional natural language understanding component.

Early approaches for this task focused on using $n$-gram models for classification \citep{reithinger1997dialogue, bird1995dialogue}. For example, Reithinger et al.\@ assumed that each dialogue act is generated by its own language model. They trained an $n$-gram language model on the utterances of each dialogue act, $P_\theta(U|A)$, and afterwards use Bayes' rule to assign the probability of a new dialogue act $P_\theta(A|U)$ to be proportional to the probability of generating the utterance under the language model $P_\theta(U|A)$.

However, a major problem with this approach is the lack of datasets with annotated dialogue acts. More recent work by \shortcite{forguesbootstrapping} acknowledged this problem, and tried to overcome the data scarcity issue by leveraging word embeddings learned from other, larger text corpora. They created an utterance-level representation by combining the word embeddings of each word, for example, by summing the word embeddings or taking the maximum w.r.t.\@ each dimension. These utterance-level representations, together with word counts, were then given as inputs to a linear classifier to classify the dialogue acts. Thus, Forgues et al.\@ showed that by leveraging another, substantially larger, corpus they were able to improve performance on their original task.

This makes the work on dialogue act classification very appealing from a data-driven perspective. First, it seems that the accuracy can be improved by leveraging alternative data sources. Second, unlike the dialogue state tracking models, dialogue act classification models typically involve relatively little feature hand-crafting thus suggesting that data-driven approaches may be more powerful for these tasks.

\subsubsection{Dialogue State Tracking}

The core task of the DSTC~\citep{williams2013dialog} adds more complexity by focusing on tracking the state of a conversation. 
This is framed as a classification problem: for every time step $t$ of the dialogue, the model is given the current input to the dialogue state tracker (including ASR and SLU outputs) together with external knowledge sources (e.g.\@ bus timetables). The required output is a probability distribution over a set of $N_t$ predefined hypotheses, in addition to the REST hypothesis (which represents the probability that none of the previous $N_t$ hypotheses are correct).
The goal is to match the distribution over hypotheses as closely as possible to the real annotated data.
By providing an open dataset with accurate labels, it has been possible for researchers to perform rigourous comparative evaluations of different classification models for dialogue systems.


Models for the DSTC include both statistical approaches and hand-crafted systems. An example of the latter is the system proposed in \shortcite{wang2013simple}, which relies on having access to a marginal confidence score $P_t(u, s, v)$ for a user dialogue $u(s = v)$ with slot $s$ and value $v$ given by a subsystem at time $t$. The marginal confidence score gives a heuristic estimate of the probability of a slot taking a particular value. The model must then aggregate all these estimates and confidence scores to compute probabilities for each hypothesis.

In this model, the SLU component may for example give the marginal confidence score (\textit{inform(data.day=today)=0.9}) in the bus scheduling DSTC, meaning that it believes with high confidence (0.9) that the user has requested information for the current day. 
This marginal confidence score 
is used to update the belief state of the system $b_t(s, v)$ at time $t$ using a set of hand-crafted updates to the probability distribution over hypotheses. From a data-driven learning perspective, this approach does not make efficient use of the dataset, but instead relies heavily on the accuracy of the hand-crafted tracker outputs.


More sophisticated models for the DSTC take a dynamic Bayesian approach by modeling the latent dialogue state and observed tracker outputs in a directed graphical model \citep{thomson2010bayesian}. These models are sometimes called generative state tracking models, though they are still discriminative in nature as they only attempt to model the state of the dialogue and not the words and speech acts in each dialogue. For simplicity we drop the index $i$ in the following equations. Similar to before, let $x_t$ be the observed tracker outputs at time $t$. Let $s_t$ be the dialogue state at time $t$, which represents the state of the world including, for example, the user actions (e.g.\@ defined by slot-value pairs) and system actions (e.g.\@ number of times a piece of information has been requested). For the DSTC, the state $s_t$ must represent the true current slot-value pair at time $t$. 
Let $r_t$ be the reward observed at time $t$, and let $a_t$ be the action taken by the dialogue system at time $t$. This general framework, also known as a partially-observable Markov decision process (POMDP) then defines the graphical model:
\begin{align}
P_{\theta}(x_t, s_t, r_t | a_t, s_{t-1}) & = P_{\theta}(x_t | s_t, a_t)
 P_{\theta}(s_t | s_{t-1}, a_t) P_{\theta}(r_t | s_t, a_t),
\end{align}
where $a_t$ is assumed to be a deterministic variable of the dialogue history. This variable is given in the DSTC, because it comes from the policy used to interact with the humans when gathering the datasets. This approach is attractive from a data-driven learning perspective, because it models the uncertainty (e.g.\@ noise and ambiguity) inherent in all variables of interest. Thus, we might expect such a model to be more robust in real applications.

Now, since all variables are observed in this task, and since the goal is to determine $s_t$ given the other variables, we are only interested in:
\begin{align}
P_{\theta}(s_t | x_t, r_t, a_t) & \propto P_{\theta}(x_t | s_t, a_t)
 P_{\theta}(s_t | s_{t-1}, a_t) P_{\theta}(r_t | s_t, a_t),
\end{align}
which can then be normalized appropriately since $s_t$ is a discrete stochastic variable. However, due to the temporal dependency between $s_t$ and $s_{t-1}$, the complexity of the model is similar to a hidden Markov model, and thus both learning and inference become intractable when the state, observation and action spaces are too large. Indeed, as noted by \shortcite{young2013pomdp}, the number of states, actions and observations can easily reach $10^{10}$ configurations in some dialogue systems. Thus, it is necessary to make simplifying assumptions on the distribution $P_{\theta}(s_t | x_t, r_t, a_t)$ and to approximate the learning and inference procedures \citep{young2013pomdp}. With appropriate structural assumptions and approximations, these models perform well compared to baseline systems on the DSTC \citep{black2011spoken}. 

Non-bayesian data-driven models have also been proposed. These models are sometimes called discriminative state tracking models, because they do not assume a generation process for the tracker outputs, $x_t$ or for any other variables, but instead only condition on them. For example, \shortcite{henderson2013deep} proposed to use a feed-forward neural network. At each time step $t$, they extracted a set of features and then concatenate a window of $W$ feature vectors together. These are given as input to the neural network, which outputs the probability of each hypothesis from the set of hypotheses. By learning a discriminative model and using a window over the last time steps, they do not face the intractability issues of dynamic Bayesian networks. Instead, their system can be trained with gradient descent methods. This approach could eventually scale to large datasets, and is therefore very attractive for data-driven learning. However, unlike the dynamic Bayesian approaches, these models do not represent probability distributions over variables apart from the state of the dialogue. Without probability distributions, it is not clear how to define a confidence interval over the predictions.  Thus the models might not provide adequate information to determine when to seek confirmation or clarification following unclear statements.

Researchers have also investigated the use of conditional random fields (CRFs) for state tracking \citep{ren2013dialog}. This class of models also falls under the umbrella of discriminative state tracking models; however, they are able to take into account temporal dependencies within dialogues by modeling a complete joint distribution over states:
\begin{align}
P_{\theta}(S | X) \propto
\prod_{c \in C} \prod_{i} f_i(\mathbf{s_c}, \mathbf{x_c}),
\end{align}
where $C$ is the set of factors, i.e.\@ sets of state and tracker variables across time, $\mathbf{s_c}$ is the set of states associated with factor $c$, $\mathbf{x_c}$ is the set of observations associated with factor $c$, and $\{f_i\}_i$ is a set of functions parametrized by parameters $\theta$. There exist certain functions $f_i$, for which exact inference is tractable and learning the parameters $\theta$ is efficient \citep{koller2009probabilistic,serban2012maximum}. For example, \shortcite{ren2013dialog} propose a set of factors which create a linear dependency structure between the dialogue states while conditioning on all the observed tracker outputs:
\begin{align}
P_{\theta}(S | X) \propto
\prod_{t} \prod_{i} f_i(\mathbf{s_{t-1}}, \mathbf{s_t}, \mathbf{s_{t+1}}, X).
\end{align}
This creates a dependency between all dialogue states, forcing them be coherent with each other. This should be contrasted to the feed-forward neural network approach, which does not enforce any sort of consistency between different predicted dialogue states. The CFR models can be trained with gradient descent to optimize the exact log-likelihood, but exact inference is typically intractable. Therefore, an approximate inference procedure, such as loopy belief propagation, is necessary to approximate the posterior distribution over states $s_t$.


In summary, there exist different approaches to building discriminative learning architectures for dialogue. While they are fairly straightforward to evaluate and often form a crucial component for real-world dialogue systems, by themselves they only offer a limited view of what we ultimately want to accomplish with dialogue models. They often require labeled data, which is often difficult to acquire on a large scale (except in the case of answer re-ranking) and require manual feature selection, which reduces their potential effectiveness. Since each model is trained independently of the other models and components with which it interacts in the complete dialogue system, one cannot give guarantees on the performance of the final dialogue system by evaluating the individual models alone. Thus, we desire models that are capable of producing probability distributions over all possible responses instead of over all annotated labels---in other words, models that can actually \textit{generate} new responses by selecting the highest probability next utterance. This is the subject of the next section.

\subsection{Response Generation Models}

Both the response re-ranking approach and the generative response model approach have allowed for the use of large-scale unannotated dialogue corpora for training dialogue systems. We therefore close this section by discussing these classes of approaches

In general, approaches which aim to generate responses have the potential to learn semantically more powerful representations of dialogues compared to models trained for dialogue state tracking or dialogue act classification tasks: the concepts they are able to represent are limited only by the content of the dataset, unlike the dialogue state tracking or dialogue act classification models which are limited by the annotation scheme used (e.g.\@ the set of possible slot-value pairs pre-specified for the DSTC).

\subsubsection{Re-ranking Response Models}

Researchers have recently turned their attention to the problem of building models that produce answers by re-ranking a set of candidate answers, and outputting the one with the highest rank or probability. While the task may seem artificial, the main advantage is that it allows the use of completely un-annotated datasets. Unlike dialogue state tracking, this task does not require datasets where experts have labeled every utterance and system response. This task only requires knowing the sequence of utterances, which can be extracted automatically from transcribed conversations.

\shortcite{banchs2012iris} construct an information retrieval system based on movie scripts using the vector space model. Their system searches through a database of movie scripts to find a dialogue similar to the current dialogue with the user, and then emits the response from the closest dialogue in the database. Similarly, \shortcite{ameixa2014luke} also use an information retrieval system, but based on movie subtitles instead of movie scripts. They show that their system gives sensible responses to questions, and that bootstrapping an existing dialogue system from movie subtitles improves answering out-of-domain questions. 
Both approaches assume that the responses given in the movie scripts and movie subtitle corpora are appropriate. Such information retrieval systems consist of a relatively small set of manually tuned parameters. For this reason, they do not require (annotated) labels and can therefore take advantage of raw data (in this case movie scripts and movie subtitles). However, these systems are effectively nearest-neighbor methods. They do not learn rich representations from dialogues which can be used, for example, to generalize to previously unseen situations. Furthermore, it is unclear how to transform such models into full dialogue agents. They are not robust and it is not clear how to maintain the dialogue state.
Contrary to search engines, which present an entire page of results, the dialogue system is only allowed to give a single response to the user.

\citep{lowe2015ubuntu} also propose a re-ranking approach using the Ubuntu Dialogue Corpus. The authors propose an \emph{affinity model} between a context $c$ (e.g.\@ five consecutive utterances in a conversation) and a potential reply $r$. Given a context-reply pair the model compares the output of a context-specific LSTM against that of a response-specific LSTM neural network and outputs whether or not the response is correct for the given context. The model maximizes the likelihood of a correct context-response pair:
\begin{align}
\max_{\theta} \sum_i P_{\theta}(\text{true response} \mid c_i, r_i)^{I_{c_i}(r_i)} 
                ( 1- P_{\theta}(\text{true response} \mid c_i, r_i))^{1-I_{c_i}(r_i)}
\end{align}
where $\theta$ stands for the set of all model parameters and $I_{c_i}(\cdot)$ denotes a function that returns 1 when $r_i$ is the correct response to $c_i$ and 0 otherwise. 
Learning in the model uses stochastic gradient descent. As is typical with neural network architectures, this learning procedure scales to large datasets. Given a context, the trained model can be used to pick an appropriate answer from a set of potential answers. This model assumes that the responses given in the corpus are appropriate (i.e., this model does not generate novel responses). However, unlike the above information retrieval systems, this model is not provided with a similarity metric as in the vector space model, but instead must learn the semantic relevance of a response to a context. This approach is more attractive from a data-driven learning perspective because it uses the dataset more efficiently and avoids costly hand tuning of parameters.

\subsubsection{Full Generative Response Models} \label{subseq:generative_response_models}

Generative dialogue response strategies are designed to automatically produce utterances by composing text (see Section~\ref{sec:DiscVsGen}). A straightforward way to define the set of dialogue system actions is by considering them as sequences of words which form utterances. \shortcite{sordoni2015aneural} and \shortcite{2015arXiv150704808S} both use this approach. They assume that both the user and the system utterances can be represented by the same generative distribution:
\begin{align}
P_{\theta}(u_1, \dots, u_T) & = \prod_{t=1}^T P_{\theta}(u_t \mid u_{<t}) \\ 
& = \prod_{t=1}^T \prod_{n=1}^N P_{\theta}(w_{t,n} \mid w_{t,<n}, u_{<t}), \label{eq:generative_model_at_word_level}
\end{align}
where the dialogue consists of $T$ utterances $u_1, \dots, u_T$ and $w_{t,n}$ is the $n^{th}$ token in utterance $t$. The variable $u_{<t}$ indicates the sequence of utterances which preceed $u_t$ and similarly for $w_{t,<n}$. Further, the probability of the first utterance is defined as $P(u_1 | u_{<1}) = P(u_1)$, and the first word of each utterance only conditions on the previous utterance, i.e.\@ $w_{t,<1}$ is ``null''. 
Tokens can be words, as well as speech and dialogue acts. The set of tokens depends on the particular application domain, but in general the set must be able to represent all desirable system actions.
In particular, the set must contain an end-of-utterance token to allow the model to express turn-taking. 
This approach is similar to language modeling. For differentiable models, training is based on maximum log-likelihood using stochastic gradient descent methods.
As discussed in Subsection \ref{sec:DiscVsGen}, these models project words and dialogue histories onto an Euclidian space. Furthermore, when trained on text only, they can be thought of as unsupervised machine learning models.


\shortcite{sordoni2015aneural} use the above approach to generate responses for posts on Twitter.
Specifically, $P_\theta(u_m\mid u_{<m})$ is given by a recurrent neural network which generates a response word-by-word based on Eq.\@ \eqref{eq:generative_model_at_word_level}. The model learns its parameters using stochastic gradient descent on a corpus of Twitter messages. The authors then combine their generative model with a machine translation system and demonstrate that the hybrid system outperforms a state-of-the-art machine translation system \citep{ritter2011data}.

\shortcite{2015arXiv150704808S} extend the above model to generate responses for movie subtitles and movie scripts. Specifically, \citet{2015arXiv150704808S} adapt a hierarchical recurrent neural network \citep{sordoni2015ahier}, which they argue is able to represent the common ground between the dialogue interlocutors. They also propose to add speech and dialogue acts to the vocabulary of the model to make the interaction with the system more natural. However, since the model is used in a standalone manner, i.e., without combining it with a machine translation system, the majority of the generated responses are highly generic (e.g.\@ \textit{I'm sorry} or \textit{I don't know}). The authors conclude that this is a limitation of all neural network-based generative models for dialogue (e.g., \citep{2015arXiv150704808S,sordoni2015aneural,vinyals2015neural}). The problem appears to lie in the distribution of words in the dialogue utterances, which primarily consist of pronouns, punctuation tokens and a few common verbs but rarely nouns, verbs and adjectives. When trained on a such a skewed distribution, the models do not learn to represent the semantic content of dialogues very well. This issue is exacerbated by the fact that dialogue is inherently ambiguous and multi-modal, which makes it more likely for the model to fall back on a generic response. As a workaround, \shortcite{li2015diversity} increase response diversity by changing the objective function at generation time to also maximize the mutual information between the context, i.e.\@ the previous utterances, and the response utterance. However, it is not clear what impact this artificial diversity has on the effectiveness or naturalness of the dialogue system. It is possible that the issue may require larger corpora to learn semantic representations of dialogue, more context (e.g.\@ longer conversations, user profiles and task-specific corpora) and multi-modal interfaces to reduce uncertainty. 
Further research is needed to resolve this question.

\shortcite{wen2015stochastic} train a neural network to generate natural language responses for a closed-dialogue domain. They use Amazon Mechanical Turk\footnote{\url{http://www.mturk.com}} to collect a dataset of dialogue acts and utterance pairs. They then train recurrent neural networks to generate a single utterance as in Eq.\@ \eqref{eq:generative_model_at_word_level}, but condition on the specified dialogue act:
\begin{align}
P_{\theta}(U | A) = \prod_{n} P_{\theta}(w_{n} \mid w_{<n}, A),
\end{align}
where $A$ is the dialogue act represented by a discrete variable, $U$ is the generated utterance given $A$ and $w_{n}$ is the $n^{th}$ word in the utterance. Based on a hybrid approach combining different recurrent neural networks for answer generation and convolutional neural networks for re-ranking answers, they are able to generate diverse utterances representing the dialogue acts in their datasets.

Similar to the models which re-rank answers, generative models may be used as complete dialogue systems or as response generation components of other dialogue systems. However, unlike the models which re-rank answers, the word-by-word generative models can generate entirely new utterances never seen before in the training set. Further, in certain models such as those cited above, response generation scales irrespective of dataset size. 



\subsection{User Simulation Models}


In the absence of large datasets, some researchers have turned to building user simulation models (sometimes referred to as `user models') to train dialogue strategies. User simulation models aim to produce natural, varied and consistent interactions from a fixed corpus, as stated by \citet[p.\@ 2]{pietquin2013survey}: ``An efficient user simulation should not only reproduce the statistical distribution of dialogue acts measured in the data but should also reproduce complete dialogue structures.'' As such, they model the conditional probability of the user utterances given previous user and system utterances:
\begin{align}
P_{\theta}(u_t^{\text{user}} | u_{<t}^{\text{user}}, u_{<t}^{\text{system}}) \label{eq:user_simulator_model},
\end{align}
where $\theta$ are the model parameters, $u_t^{\text{user}}$ and $u_t^{\text{system}}$ are the user utterance (or action) and the system utterance (or action) respectively at time $t$. Similarly, $u_{<t}^{\text{user}}$ and $u_{<t}^{\text{system}}$ indicate the sequence of user and system utterances that precede $u_t^{\text{user}}$ and $u_t^{\text{system}}$, respectively. 

There are two main differences between user simulation models and the generative response models discussed in Subsection \ref{subseq:generative_response_models}. First, user simulation models never model the distribution over system utterances, but instead only model the conditional distribution over user utterances given previous user and system utterances. Second, user simulation models usually model dialogue acts as opposed to word tokens. Since a single dialogue act may represent many different utterances, the models generalize well across paraphrases. However, training such user simulation models requires access to a dialogue corpus with annotated dialogue acts, and limits their application to training dialogue systems which work on the same set of dialogue acts. For spoken dialogue systems, user simulation models are usually combined with a model over speech recognition errors based on the automatic speech recognition system but, for simplicity, we omit this aspect in our analysis.

Researchers initially experimented with $n$-gram-based user simulation models \citep{eckert1997user,georgila2006user}, which are defined as:
\begin{align}
P_{\theta}(u_t^{\text{user}} | u_{t-1}^{\text{system}}, u_{t-2}^{\text{user}}, \dots, u_{t-n-1}^{\text{system}}) \label{eq:user_simulator_model_n_gram} = \theta_{u_t^{\text{user}}, u_{t-1}^{\text{system}}, u_{t-2}^{\text{user}}, \dots, u_{t-n-1}^{\text{system}}},
\end{align}
where $n$ is an even integer, and $\theta$ is an $n$-dimensional tensor (table) which satisfies:
\begin{align}
\sum_{u_t^{\text{user}}} \theta_{u_t^{\text{user}}, u_{t-1}^{\text{system}}, u_{t-2}^{\text{user}}, \dots, u_{t-n-1}^{\text{system}}} = 1.
\end{align}
These models are trained either to maximize the log-likelihood of the observations by setting 
$\theta_{u_t^{\text{user}}, u_{t-1}^{\text{system}}, u_{t-2}^{\text{user}}, \dots, u_{t-n-1}^{\text{system}}}$ equal to (a constant times) the number of occurrences of each corresponding  $n$-gram 
, or on a related objective function which encourages smoothness and therefore reduces data sparsity for larger $n$'s \citep{joshua2001bit}. Even with smoothing, $n$ has to be kept small and these models are therefore unable to maintain the history and goals of the user over several utterances \citep{schatzmann2005quantitative}. Consequently, the goal of the user changes over time, which has a detrimental effect on the performance of the dialogue system trained using the user simulator.

Several solutions have been proposed to solve the problem of maintaining the history of the dialogue. \citet{pietquin2004framework} propose to condition the $n$-gram model on the user's goal:
\begin{align}
P_{\theta}(u_t^{\text{user}} | u_{t-1}^{\text{system}}, u_{t-2}^{\text{user}}, \dots, u_{t-n-1}^{\text{system}}, g),
\end{align}
where $g$ is the goal of the user defined as a set of slot-value pairs. Unfortunately, not only must the goal lie within a set of hand-crafted slot-value pairs, but its distribution when simulating must also be defined by experts. 
Using a more data-driven approach, \citet{georgila2006user} propose to condition the $n$-gram model on additional features:
\begin{align}
P_{\theta}(u_t^{\text{user}} | u_{t-1}^{\text{system}}, u_{t-2}^{\text{user}}, \dots, u_{t-n-1}^{\text{system}}, f(u_{<t}^{\text{user}}, u_{<t}^{\text{system}})),
\end{align}
where $f(u_{<t}^{\text{user}}, u_{<t}^{\text{system}})$ is a function mapping all previous user and system utterances to a low-dimensional vector that summarizes the previous interactions between the user and the system (e.g.\@ slot-value pairs that the user has provided the system up to time $t$). Now, $\theta$ can be learned using maximum log-likelihood with stochastic gradient descent.

More sophisticated probabilistic models have been proposed based on directed graphical models, such as hidden Markov models and input-output hidden Markov models \citep{cuayahuitl2005human}, and undirected graphical models, such as conditional random fields based on linear chains \citep{jung2009data}. Inspired by \citet{pietquin2005probabilistic}, \citet{pietquin2007learning} and \citet{rossignol2011training} propose the following directed graphical model:
\begin{align}
P_{\theta}(u_t^{\text{user}} | u_{<t}^{\text{user}}, u_{<t}^{\text{system}}) = \sum_{g_t, k_t} P_{\theta}(u_t^{\text{user}} | g_t, k_t, u_{<t}^{\text{user}}, u_{<t}^{\text{system}}) P_{\theta}(g_t | k_t) P_{\theta}(k_t | k_{<t}, u_{<t}^{\text{user}}, u_{<t}^{\text{system}})
\end{align}
where $g_t$ is a discrete random variable representing the user's goal at time $t$ (e.g.\@ a set of slot-value pairs), and $k_t$ is another discrete random variable representing the user's knowledge at time $t$ (e.g.\@ a set of slot-value pairs). This model allows the user to change goals during the dialogue, which would be the case, for example, if the user is notified by the dialogue system that the original goal cannot be accomplished. The dependency on previous user and system utterances for $u_t^{\text{user}}$ and $k_t$ may be limited to a small number of previous turns as well as a set of hand-crafted features computed on these utterances. For example, the conditional probability:
\begin{align}
P_{\theta}(u_t^{\text{user}} | g_t, k_t, u_{<t}^{\text{user}}, u_{<t}^{\text{system}}),
\end{align}
may be approximated by an $n$-gram model with additional features as in \citet{georgila2006user}. Generating user utterances can be done in a straightforward manner by using ancestral sampling: first, sample $k_t$ given $k_{<t}$ and the previous user and system utterances; then, sample $g_t$ given $k_t$; and finally, sample $u_t^{\text{user}}$ given $g_t$, $k_t$ and the previous user and system utterances. The model can be trained using maximum log-likelihood. If all variables are observed, i.e.\@ $g_t$ and $k_t$ have been given by human annotators, then the maximum-likelihood parameters can be found similarly to $n$-gram models by counting the co-occurrences of variables. If some variables are missing, they can be estimated using the expectation-maximization (EM) algorithm, since the dependencies form a linear chain. \citet{rossignol2011training} also propose to regularize the model by assuming a Dirichlet distribution prior over the parameters, which is straightforward to combine with the EM algorithm.

User simulation models are particularly useful in the development of dialogue systems based on reinforcement learning methods~\citep{singh2002optimizing,schatzmann2006survey,pietquin2006probabilistic,frampton2009recent,jurvcivcek2012reinforcement,png2011bayesian,young2013pomdp}. Furthermore, many user simulation models, such as those trainable with stochastic gradient descent or co-occurrence statistics, are able to scale to large corpora. In the light of the increasing availability of large dialogue corpora, there are ample opportunities for building novel user simulation models, which aim to better represent real user behavior, and in turn for training dialogue systems, which aim to solve more general and more difficult tasks. Despite their similarities, research on user simulation models and full generative models has progressed independently of each other so far. Therefore, it also seems likely that there is fruitful work to be done in transferring and merging ideas between these two areas.

\end{document}